\documentclass{article}

\usepackage{arxiv}

\usepackage[utf8]{inputenc} 
\usepackage[T1]{fontenc}    
\usepackage{hyperref}       
\usepackage{url}            
\usepackage{booktabs}       
\usepackage{amsfonts}       
\usepackage{nicefrac}       
\usepackage{microtype}      
\usepackage{lipsum}		
\usepackage{graphicx}
\usepackage{natbib}
\usepackage{doi}

\usepackage{amsmath}
\usepackage{setspace}
\usepackage{algpseudocode}
\usepackage{algorithm}
\usepackage{tabularx,tabulary}
\usepackage{setspace}
\usepackage{gensymb}
\usepackage{graphicx}
\usepackage{xcolor}
\usepackage{colortbl}
\usepackage{multirow}

\DeclareUnicodeCharacter{2212}{-}

\title{Enhancing Fairness and Performance in Prediction Models: A Multi-Task Learning Approach with Monte-Carlo Dropout and Pareto Optimality}

\author{ \href{https://orcid.org/0000-0002-3553-303X}{\includegraphics[scale=0.06]{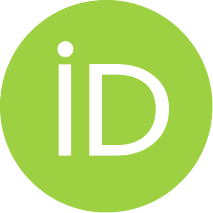}\hspace{1mm}Khadija Zanna}\thanks{Use footnote for providing further
		information about author (webpage, alternative
		address)---\emph{not} for acknowledging funding agencies.} \\
	Department of Electrical and Computer Engineering\\
	Rice University\\
	Houston, TX 77005 \\
	\texttt{khzanna@rice.edu} \\
	\And
	\href{https://orcid.org/0000-0003-4484-8946}{\includegraphics[scale=0.06]{orcid.pdf}\hspace{1mm}Akane Sano} \\
	Department of Electrical and Computer Engineering\\
	Rice University\\
	Houston, TX 77005 \\
	\texttt{Akane.Sano@rice.edu} \\
}

\renewcommand{\shorttitle}{Enhancing Fairness and Performance in Prediction
Models}

\hypersetup{
pdftitle={A template for the arxiv style},
pdfsubject={q-bio.NC, q-bio.QM},
pdfauthor={David S.~Hippocampus, Elias D.~Striatum},
pdfkeywords={First keyword, Second keyword, More},
}

\begin{document}
\maketitle

\begin{abstract}
Bias originates from both data and algorithmic design, often exacerbated by traditional fairness methods that fail to address the subtle impacts of protected attributes. This study introduces an approach to mitigate bias in machine learning by leveraging model uncertainty. Our approach utilizes a multi-task learning (MTL) framework combined with Monte Carlo (MC) Dropout to assess and mitigate uncertainty in predictions related to protected labels. By incorporating MC Dropout, our framework quantifies prediction uncertainty, which is crucial in areas with vague decision boundaries, thereby enhancing model fairness. Our methodology integrates multi-objective learning through pareto-optimality to balance fairness and performance across various applications. We demonstrate the effectiveness and transferability of our approach across multiple datasets and enhance model explainability through saliency maps to interpret how input features influence predictions, thereby enhancing the interpretability of machine learning models in practical applications.
\end{abstract}


\section{Introduction}
\label{Introduction}

Introduced by Tom Mitchell in 1980 (\cite{mitchell1980need}), the term "bias" in machine learning refers to giving importance to particular features to improve generalization—which is essential for model performance. On the contrary, bias can also be negative, leading to inaccurate assumptions by the algorithm that can be prejudiced against certain groups of people (\cite{zanna2022bias}). These decisions could cause adverse effects on particular social groups, for example, those defined by sex, race, age, marital status, handicaps, etc., when used to make autonomous decisions in life-changing cases such as health, hiring, education, criminal sentencing, etc.

Negative bias arises either from the data itself or the algorithm's design (\cite{blanzeisky2021algorithmic}). Data-related biases, or 'negative legacies' (\cite{kamishima2012fairness}), stem from unrepresentative training samples or labeling errors, while algorithmic biases or "underestimation" occurs when the classifier under-predicts an uncommon outcome for the minority group. This can happen when the classifier under-fits the data, causing it to focus on strong signals and missing more subtle phenomena (\cite{cunningham2021underestimation}).

In recent years, research on bias and fairness in machine learning has shown that simply eliminating the sensitive features from the model's workings is insufficient to avoid an unfavorable determination process. This is due to the indirect influence of sensitive information on other features vital to the algorithm's decision-making process. As mentioned by \cite{paulus2020predictably}, the use of “high dimensional” or “black box” prediction techniques is generally problematic given that these approaches can predict the protected label through other variables, whether or not the actual protected label is explicitly encoded. Given these scenarios, it is understood that reducing model bias may not be sufficient in eliminating fairness concerns in decision contexts associated with certain fields such as predictive healthcare, which can be complex, unintuitive, and often hard to explain (\cite{yang2022explainable}). This highlights the need for bias solutions to strive for better explainability to improve understanding of the model, increase the utility of its output, and produce better outcomes (\cite{yang2022explainable}).

Another setback with developing methods to mitigate bias is the trade-off between performance and fairness, which is prevalent in existing literature (\cite{pessach2022review,buijsman2023navigating}). As a solution to this issue, several previous works have used the concept of Pareto optimality from multi-objective optimization to seek the fairness-accuracy Pareto front of a neural network classifier (\cite{wei2022fairness, liu2022accuracy, wang2021understanding, liang2022algorithmic, little2022fairness, kamani2021pareto}). Multi-objective optimization addresses the problem of optimizing a set of possibly contrasting objectives (\cite{sener2018multi}). Before the introduction of this concept to fairness literature within the past 4 years, existing algorithmic fairness methods have typically performed the so-called linear scalarization scheme. This is a multi-objective optimization method that converts multiple objectives into a single composite objective (\cite{marler2004survey}). It can miss significant trade-offs between objectives because it assumes a linear relationship among them, which presents limitations in recovering Pareto optimal solutions (\cite{wei2022fairness}). 

As machine learning plays a more significant role in decision-making across industries, the demand for domain-agnostic and easily transferable bias mitigation techniques is increasing. Many machine learning algorithms are unique and proprietary, each facing different challenges related to bias. Current mitigation solutions often target specific problems and are not flexible enough to address varying fairness and performance concerns (\cite{pagano2022bias}). This situation highlights the need for versatile bias mitigation methods suitable for diverse domains.

Our research introduces a bias mitigation approach that utilizes model uncertainty to enhance fairness in machine learning. Our method leverages epistemic uncertainty which arises from incomplete knowledge about model parameters and can be reduced with additional data (\cite{stone2022epistemic}). This uncertainty correlates with model weights and highlights biases which can be addressed through mitigation strategies (\cite{li2021deep}). Specifically, using weights from models uncertain about a protected label helps diminish its influence on the outcome by averaging over multiple plausible models.

To evaluate the uncertainty of the protected label, we employ a multi-task learning (MTL) framework to predict target and protected labels while assessing their associated uncertainties. This framework enhances learning by sharing information across tasks, improving model efficacy (\cite{caruana1997multitask}). To model uncertainty, we employ Monte Carlo (MC) Dropout to perform multiple forward passes with varied dropout masks, creating a distribution of outputs to estimate uncertainty accurately (\cite{milanes2021monte}). The ensemble approach of MC Dropout during inference maintains high performance by reducing the impact of random dropout noise in individual predictions (\cite{ahmed2023scale}). This technique also quantifies prediction uncertainty, essential in regions with vague decision boundaries or insufficient data (\cite{choubineh2023applying}), which often disproportionately affects marginalized groups, underscoring the importance of fairness (\cite{kendall2017uncertainties}). By focusing adjustments on areas of highest uncertainty, especially in sensitive applications, our method enhances fairness and model robustness.

Extending our prior work (\cite{zanna2022bias}), we also integrate multi-objective learning via pareto-optimality into this framework, allowing us to develop models that achieve the optimal balance between fairness and performance. This flexibility ensures that the results can be tailored to meet specific needs across various domains, where one metric might be prioritized over others. We further analyze and test this framework on multiple classification-based datasets, including two widely-used public datasets (ADULT \& MIMIC-III) and a private dataset (SNAPSHOT) across diverse domains such as in-hospital mortality, finance, and stress prediction. This demonstrates the method's transferability across different fields. Notably, the ADULT dataset is extensively utilized in bias studies, highlighting its significance in this research.

Furthermore, our approach enhances model explainability without incorporating sensitive data directly. By employing a saliency map technique (\cite{simonyan2013deep}), we interpret how input features influence predictions and highlight our method's impact on feature prioritization. This feature will allow practitioners in various fields to interpret better the decision-making process of the models they utilize. The major contributions of this work include: 

\begin{itemize}
    \item Introduced a bias mitigation approach that leverages MTL and MC Dropout which operates independently of the machine learning model's objective function and prediction methodology, avoiding the use of sensitive data in the model’s functional form.
    \item Integrated Multi-Objective Learning Through Pareto-Optimality to balance fairness and performance, tailoring outcomes to specific domain needs where certain metrics may be prioritized.
    \item Validated the proposed method on several classification-based datasets, including the ADULT and MIMIC-III public datasets, and a non-public SNAPSHOT dataset from three domains. 
    \item Compared performance of the proposed method to other baseline methods namely: Reweighting, ARL, and FairRF.
    \item Employed a saliency map technique to visualize the significance of model weights, improving our understanding of how input features influence final predictions. This visualization also highlights the impact of our proposed method on the network's prioritization of features for prediction.
\end{itemize}

\section{Related Works}
\label{related_works}

\subsection{Fairness Challenges in Machine Learning}

Fairness has long been studied in philosophy and psychology and has recently become critical in machine learning (\cite{pleiss2017fairness}). Generally, fairness can be defined as the absence of prejudice or favoritism towards an individual or a group based on their intrinsic or acquired traits in decision-making (\cite{saxena2019fairness}). Even though the necessity of fairness has been demonstrated through numerous perspectives, it can be challenging to achieve in practice in machine learning and society as a whole (\cite{mehrabi2021survey}). There are several bias mitigation techniques (\cite{calders2010three, chouldechova2017fair, kamiran2012data, zafar2017fairness, zhang2018mitigating}) introduced in prior literature, which are generally categorized into 3 groups: pre-processing, in-processing, and post-processing methods. 

Pre-processing techniques are used if the algorithm is allowed to modify the training data. They are generally easier to implement, and there is no need to access sensitive labels at test time, which is not allowed in some jurisdictions (\cite{d2017conscientious, yan2020mitigating}). These methods improve fairness, but they are very domain-oriented and could pose some domain-related limitations where for example with health-related data, generation of synthetic data can be tricky (\cite{chen2021synthetic}). Some works utilize the reweighing pre-processing technique for bias mitigation (\cite{paviglianiti2020vital, vega2022fair, krasanakis2018adaptive}). This method has shown to work reasonably well in improving fairness and has become a standard in the field, but it usually struggles when the dataset does not contain sufficient samples for a specific demographic group (\cite{yang2022enhancing}). Our proposed method which is a hybrid between in-processing and post-processing does not incur the domain-related limitations that pre-processing techniques typically suffer from. 

In-processing techniques modify learning algorithms to eliminate discrimination during model training by altering the objective function or adding constraints (\cite{gorrostieta2019gender, liu2022accuracy}). These techniques are applicable when changes to the learning process are permissible, and sensitive attribute data is available (\cite{yan2020mitigating, gorrostieta2019gender}). In recent years, in-processing techniques have gained popularity due to their enhanced accuracy and fairness (\cite{pagano2022bias}). However, these methods are typically task-specific and non-transferable because they require direct modifications to the model, which may not always be feasible and can incur high performance costs (\cite{oneto2019taking}). Our proposed method overcomes these limitations by utilizing Pareto optimality, allowing for flexible tuning between fairness and performance. Additionally, it does not necessitate changes to the machine learning model's core functions, thus eliminating the need for task-specific alterations commonly associated with in-processing techniques.

Post-processing bias mitigation methods aim to edit posteriors in ways that satisfy the fairness constraint, and no changes are made to the model (\cite{oneto2019taking}). These techniques are suitable when a machine learning model must be treated as a black box, without modifications to training data or algorithms. They are applicable to any classifier and generally perform well in fairness.  However, they lack flexibility in balancing accuracy with fairness (\cite{oneto2019taking, yan2020mitigating}), an issue our method addresses using Pareto optimality. Current bias mitigation methods also offer limited support for black-box models, unlike the extensive resources available for white-box models.

\subsection{Multi-task Learning for Bias Mitigation}

Multi-task learning is a machine learning technique where a model is trained to perform multiple tasks simultaneously. It aims to leverage useful information contained in multiple related tasks to help improve the generalization performance of all the tasks (\cite{zhang2018overview}). By doing so, MTL not only optimizes the accuracy of each task but also enhances fairness by ensuring that the model does not disproportionately learn from or favor any single task or demographic group. Studies have demonstrated that models trained using MTL not only perform better on average but also show improved fairness metrics compared to models trained on individual tasks separately (\cite{li2023multi}).

One of the most common methods in MTL is hard parameter sharing, which involves sharing hidden layers between all tasks while keeping several task-specific output layers (\cite{li2023multi}). It has been found effective in reducing model complexity and helping prevent overfitting, which is crucial for maintaining the fairness of predictions across different demographic groups as shown in \cite{li2023multi} where different sub-populations are divided into separate tasks and the gradients of various prediction tasks during neural network back-propagation are dynamically modified to ensure fairness. This approach was tested on data from sepsis patients and was shown to significantly improve fairness disparity on the test data. 

In the 2022 publication \cite{gao10mitigate}, propose a novel approach to addressing gender bias in machine learning models through the framework of negative multi-task learning (NMTL). This technique is designed to actively reduce bias by learning tasks that negatively correlate with the biased outcomes, thereby mitigating undesirable bias propagation in predictive models. The concept of NMTL involves training models on tasks that are inversely related to biased decision-making processes, effectively using these negative correlations to counteract bias. The study presents results that demonstrate significant improvements in fairness metrics in the context of gender bias across various datasets.

Some studies have also integrated fairness directly into the MTL framework by treating fairness as an auxiliary task. For example, \cite{oneto2019taking} proposes a framework where fairness is incorporated as an auxiliary objective alongside the primary classification tasks. This allows the model to leverage shared information across tasks to reduce bias inherently present in single-task learning environments. The authors showed that their approach improved fairness metrics without a substantial loss in accuracy through experiments on synthetic and real-world datasets. 

While there are works thatlay a significant foundation in using MTL for fair classification (\cite{oneto2019taking, gao10mitigate, li2023multi}), our proposed MTL approach introduces several key improvements by integrating uncertainty estimation techniques, specifically MC Dropout, to further refine the understanding of how different input features influence the model's predictions. This approach not only helps in identifying bias but also in dynamically adjusting the model’s reliance on various features based on their contribution to unfairness. We also extend this concept of MTL by incorporating Pareto optimality principles, which allows our model to explore a broader spectrum of potential solutions that balance multiple fairness metrics and accuracy more dynamically. This is particularly useful in complex real-world scenarios where different fairness considerations might conflict. This ensures that our proposed approach is suitable for a wider range of applications due to this adaptability feature.

\subsection{Model Uncertainty for Bias Mitigation}

The utilization of model uncertainty to mitigate biases in machine learning has been explored in several recent studies, highlighting different approaches.

In their work, \cite{stone2022epistemic} introduce a method that incorporates epistemic uncertainty which refers to uncertainty caused by a lack of knowledge of the loss function used for training visual models (\cite{hullermeier2021aleatoric}). Their approach uses uncertainty measurements to weigh the loss function, prioritizing learning from less certain instances which are hypothesized to be less biased. This method helps reduce bias in visual recognition tasks by modifying how the model learns from data points with high uncertainty (\cite{stone2022epistemic}). Our approach borrows from this idea by utilizing MC Dropout to determine model uncertainty in our prediction models.

Another approach is also introduced by \cite{heuss2023predictive} where model uncertainty is applied to mitigate bias in ranking algorithms. They propose a framework where predictive uncertainty is used to adjust the rankings to be fairer across different groups, ensuring that the uncertainty in model predictions does not disproportionately affect any single group (\cite{heuss2023predictive}). \cite{almeida2021mitigating} on the other hand argued that uncertainty in class boundaries for classification tasks can lead to biased and inconsistent decisions. Their method reduces this specific type of uncertainty to improve both the fairness and the accuracy of the model.

Based on the evidence from prior works (\cite{stone2022epistemic, heuss2023predictive, almeida2021mitigating}), our research integrates uncertainty into our proposed MTL framework to enhance bias mitigation across multiple domains.

\section{Fairness Terminology}\label{sec:definitions}

In this section, we introduce and define some of the concepts and terminologies generally used in algorithmic fairness research that are relevant to this paper.

\begin{itemize}
\item \textbf{Protected (sensitive) label}: An attribute that partitions a population into groups whose outcomes should have parity (such as race, gender, income, etc.).
\item \textbf{Privileged class}: A protected label value indicating a group that is at an advantage. Generally represented by the value 1.
\item \textbf{Unprivileged class}: A protected label value indicating a group that is at a disadvantage. Generally represented by the value 0.
\item \textbf{Disparate impact ratio (DIR)}: This is a fairness metric which compares the rate of positive outcomes in the unprivileged class, with the rate of positive outcomes in the privileged class, as shown in Equation \ref{eq:disp_imp}. It is the measure of how different outcomes are for different groups, based on the results of a model (\cite{feldman2015certifying}). A value of less than 1 indicates less favorable outcomes for the unprivileged group.
In this paper, we assume an acceptable lower bound of 0.8, and an upper bound of 1.2, with 1 being the ideal score. 

\begin{equation}
   \text{DIR =} \frac{Pr(Y=1|D=unprivileged)} {Pr(Y=1|D=privileged)}
\label{eq:disp_imp}
\end{equation}

where Y represents the target label and  D the class of the protected demographic label.

\item \textbf{Equalized odds}: This fairness metric enforces that the model correctly identifies the positive outcome at equal rates across groups, and miss-classify the positive outcome at equal rates across groups (creating the same proportion of True Positives and False Positives across groups). In this work, we represent this metric using 2 values; Difference in False Negative (FN) scores, and Difference in False Positive (FP) scores.

\end{itemize}

\section{Multi-task Learning Based Bias Mitigation Method (Proposed Method)}

\subsection{Overview}

\begin{figure}[htbp]
\centerline{\includegraphics[width=\textwidth]{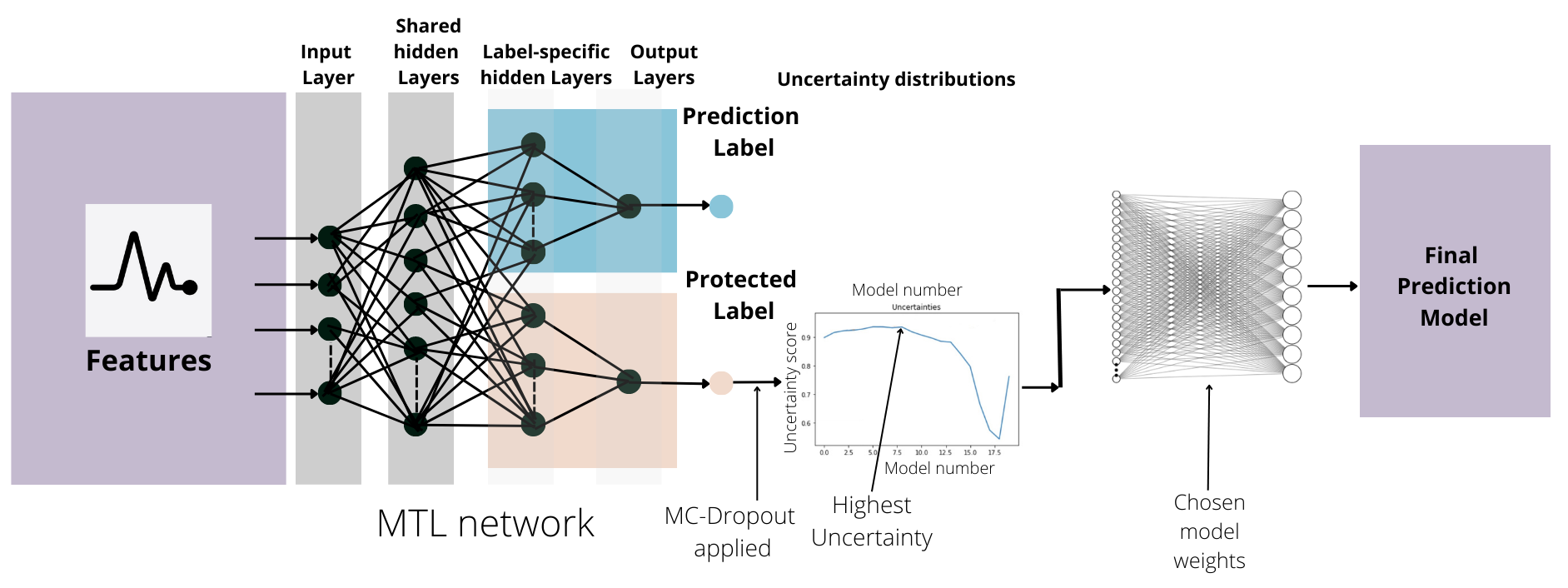}}
\caption{Proposed Method Structure}
\label{fig-method_structure}
\end{figure}

Our proposed methodology is based on the premise of Bayesian uncertainty estimation to address fairness in machine learning models. Uncertainty estimation refers to estimating the confidence level of a model output. In this case, the posterior variance represents the total uncertainty of the prediction (\cite{wang2021hybrid}). In our proposed method, we hypothesize that when the model exhibits the greatest range of uncertainty regarding the protected label, it indicates a deficiency in knowledge within specific areas of the feature space linked to that label. Utilizing the model weights from this range of uncertainty for predicting the target label is likely to diminish the influence of the protected label on the final prediction outcome. This approach is based on our findings that a spectrum of model weights, rather than a single optimal model, yields the best fairness scores. Therefore, we employ non-dominated sorting to identify the Pareto optimal set of solutions. This method will allow us to select models that effectively balance fairness and performance by recognizing that improvements in one metric may lead to reductions in the other, thus providing a robust framework for addressing fairness. Figure \ref{fig-method_structure}) represents a visual diagram of our proposed method, the following subsections detail the flow of our method with the different concepts involved.

\subsection{Multi-Task Learning Framework}

To accurately assess the uncertainty associated with both the target and protected labels, we first implement an MTL framework that optimizes the learning of correlated tasks by sharing common features while allowing for task-specific adaptations. This framework uses a shared input layer but distinct output layers for each task—one for the primary target prediction and another for protected label prediction. This framework is rooted in the principle that related tasks can share useful information and thus be learned more effectively together than separately (\cite{caruana1997multitask}).

\subsection{Monte Carlo Dropout for Uncertainty Estimation}

Our MTL model integrates Monte Carlo Dropout for uncertainty estimation. Unlike traditional dropout techniques used solely during training to combat overfitting, MC Dropout extends this concept to inference. We incorporate MC Dropout in our MTL model by performing multiple forward passes with randomly applied dropout. This simulates drawing samples from the model's posterior weight distribution. This process effectively turns our neural network into a Bayesian model, where each dropout configuration leads to different sets of predictions. From this, we get predictions for both the target and the selected protected label. Analyzing the variance across these predictions provides a robust measure of uncertainty, particularly highlighting where the model's predictions are least confident, potentially signaling underlying biases.

Modeling uncertainty with MC Dropout works by running multiple forward passes through the model with a different dropout mask every time. It draws weights from approximate posterior distributions. The posterior distribution is a way to summarize what we know about uncertain quantities in Bayesian analysis. It combines the prior distribution and the likelihood function, which tells you what information is in your observed data (the “new evidence”). In other words, the posterior distribution summarizes what we know after the data is observed. 

Given a trained neural network model with dropout \(f_{nn}\), to derive uncertainty for one sample \(x\), the Monte Carlo algorithm collects the predictions of T inferences with different dropout masks. Here, \(f_{nn}^{d_i}\) represents the model with dropout mask \(d_i\). So a sample of the possible model outputs for sample \(x\) is obtained as  \(f_{nn} ^{d_0} (x),...,f_{nn} ^{d_T} (x)\). By computing the average and variance of this sample, we obtain an ensemble prediction, which is the mean of the model's posterior distribution for this sample and an estimate of the uncertainty of the model regarding \(x\).

\begin{equation}
\label{eq:eq-mean}
    predictive\:posterior\:mean:  p = \frac{1}{T} \sum_{i=1}^{T} f_{nn} ^{d_i} (x)
\end{equation}

\begin{equation}
\label{eq:uncert}
    \text{uncertainty: } c = \frac{1}{T} \sum_{i=1}^{T}[f_{nn} ^{d_i} (x) - p]^{2}
\end{equation}

Further details on the theoretical underpinnings of MC Dropout can be found in the seminal work by \cite{gal2016dropout}. The uncertainties derived from the Monte Carlo simulations are critical for assessing model fairness. 

\subsection{Non-Dominated Sorting}
\label{ssec:nd_sorting}


From the saved models obtained using Monte-Carlo Dropout, we compute the uncertainty scores and extract the parameters from each model. These parameters are then employed in a distinct (non-multi-task) model dedicated to predicting our main target label. We train these models, monitor their performance metrics, and compute fairness metrics, including DIRs, differences in FN and FP rates from their predictions.

To identify the most effective models in terms of balancing performance and fairness, we apply a non-dominated sorting algorithm to the ensemble’s performance and fairness scores, particularly focusing on the DIRs. This process allows us to derive the Pareto optimal set of models, collectively referred to as the "Pareto Front." Models on this front represent the optimal trade-offs between performance and fairness; any improvements in one metric would necessitate a compromise in the other. This front is significant as it contains no models that outperform others across both metrics simultaneously, illustrating the inherent trade-offs between achieving high performance and maintaining fairness.

The non-dominated sorting algorithm operates by comparing models based on their performance scores (X) and fairness scores (Y - specifically DIR scores), as outlined in Algorithm \ref{alg:pareto_frontier_diff}. It ranks models that lie on the Pareto frontier, sorted in order of their optimality. The algorithm’s objective is to maximize the performance scores (X) while minimizing the absolute difference between the fairness scores (Y) and the ideal value of 1. It identifies Pareto optimal points from two metric arrays $Xs$ and $Ys$, which represent the conflicting objectives of performance and fairness. The algorithm pairs these metrics, sorts them based on the $Xs$ values according to the preference for maximization or minimization, and iteratively constructs a Pareto frontier. Each subsequent pair of X and Y is evaluated against the last point on the frontier: it is added if it offers an improvement in $Y$ without worsening $X$, measured by the absolute difference from an ideal value. The outcome is two arrays, $p\_frontX$ and $p\_frontY$, depicting the optimal trade-offs between the evaluated metrics. 

\begin{singlespace}
\begin{algorithm}
\caption{Calculate Pareto Frontier}
\label{alg:pareto_frontier_diff}
\begin{algorithmic}[1] 
\Procedure{ParetoFrontier}{$Xs, Ys, maxX, maxY$}
    \State \textbf{Input:} Metric arrays: Performance scores ($Xs$), Fairness scores ($Ys$) of equal length; booleans $maxX$, $maxY$
    \State \textbf{Output:} Arrays $p\_frontX$, $p\_frontY$ of Pareto optimal points
    \State
    \State Combine $Xs$ and $Ys$ into list $myList$ of tuples $(x, y)$
    \If{$maxX$ is True}
        \State Sort $myList$ by $x$ in descending order
    \Else
        \State Sort $myList$ by $x$ in ascending order
    \EndIf
    \State Initialize $p\_front$ with the first tuple from $myList$
    \For{each tuple $(currentX, currentY)$ in $myList$}
        \If{$maxY$}
            \If{$\text{abs}(1 - currentY) < \text{abs}(1 - y$ of last tuple in $p\_front)$}
                \State Append $(currentX, currentY)$ to $p\_front$
            \EndIf
        \Else
            \If{$\text{abs}(1 - currentY) > \text{abs}(1 - y$ of last tuple in $p\_front)$}
                \State Append $(currentX, currentY)$ to $p\_front$
            \EndIf
        \EndIf
    \EndFor
    \State Separate $p\_front$ into $p\_frontX$ and $p\_frontY$
    \State \Return $(p\_frontX, p\_frontY)$
\EndProcedure
\end{algorithmic}
\end{algorithm}
\end{singlespace}

\section{Experiments}

\subsection{Datasets}


We evaluate the effectiveness of our proposed method using three datasets, (1) ADULT (individual income and demographic information) (\cite{kohavi1996scaling}), (2) MIMIC-III (data from patients in critical care units) \cite{johnson2016mimic}, and (3) SNAPSHOT datasets (multi-modal physiological, behavioral, and survey data collected from 350 college students) (\cite{sano2015recognizing}). Table \ref{data_variables} describes the input and binary target variables for each of these datasets used in our experiments. More information on the details of these datasets and data processing can be found in Appendix A. 

\begin{singlespace}
\begin{table}[htbp]
\begin{center}
\caption{Input and Target Variables for ADULT, MIMIC-III, and SNAPSHOT Datasets}
\label{data_variables}
\begin{tabular}{ |c|c|c| }
\hline
Dataset & Input Variables & Target Variables \\
\hline \hline
\multirow{3}{6em}{ADULT} & Age, workclass, education, & Salary \\ 
& Marital-status, occupation, & \\ 
& relationship, race, sex &  \\ 
\hline

\multirow{10}{6em}{MIMIC-III} & Vital signs, medications, & In-hospital mortality (IHM) \\ 
& laboratory measurements,  &  \\ 
& fluid balance, procedure codes, &  \\
& hospital length of stay, &  \\
& survival data,  &  \\
& laboratory test results, &  \\
& diagnostic codes,  & \\
& patient demographics, & \\
& billing information, & \\
& observations and notes, etc. & \\
\hline

\multirow{4}{6em}{SNAPSHOT} & Physiological features, & Morning happiness \\ 
& mobile phone usage, & Evening happiness \\ 
& weather data & Morning Calmness \\ 
& & Evening Calmness \\
\hline

\end{tabular}
\end{center}
\end{table}
\end{singlespace}

For the ADULT dataset, we regard age, sex(gender) and race as the protected labels, and for MIMIC-III, age, gender, insurance type, and marital status. For the SNAPSHOT dataset, we assign gender, race, and ethnicity as the protected labels in our experiments. We also use personality types, including openness, conscientiousness, extraversion, agreeableness, and neuroticism, as protected labels in our experiments for SNAPSHOT. Table \ref{adult_mimic_binarize} details how we binary-encode each protected label and assign them as privileged and unprivileged classes to analyze them using our chosen fairness metrics. 

\begin{singlespace}
\begin{table}[htbp]
\centering
\caption{Binarization of Protected labels for ADULT, MIMIC-III, and SNAPSHOT Datasets}
\begin{tabular}{|l|l|l|l|}
\hline
\textbf{Dataset} & \textbf{Label} & \textbf{Privileged Class (1)} & \textbf{Unprivileged Class (0)} \\ 
\hline \hline
\multirow{3}{4em}{ADULT} & Age & Less than or equal to 40 years & Greater than 40 years \\ 
& Race & White, Asian-Pac-Islander & Black, Amer-Indian-Eskimo, other \\ 
& Sex & Male & Female \\ \hline

\multirow{3}{4em}{MIMIC-III} & Age & Greater than 60 years & Less than or equal to 60 years \\ 
& Gender & Male & Female \\ 
& Insurance & Private, other & Medicare, Medicaid \\ 
& Marital Status  & Married, Life partner & Single, Widowed, Divorced, Separated \\ \hline

\multirow{3}{4em}{SNAP-SHOT} & Gender & Male & Female \\
& Race & White & Non-white \\
& Ethnicity & Non-Hispanic/latino & Hispanic/latino  \\
& Openness & $<=50$ & $>50$ \\
& Conscientiousness & $>50$ & $<=50$ \\
& Extraversion & $<=50$ & $>50$ \\
& Agreeableness & $<=50$ & $>50$ \\
& Neuroticism & $<=50$ & $>50$ \\ \hline

\end{tabular}
\label{adult_mimic_binarize}
\end{table}
\end{singlespace}

\subsection{Baseline Models and Fairness Analyses} \label{sec:base_models}

The first part of our experiments involves developing baseline models that make the target predictions for all datasets (income for ADULT, IHM for MIMIC-III, and 4 happiness and calmness labels for SNAPSHOT). Next, we analyze our prediction results for fairness by calculating the DIR and equalized odds (FP and FN rates) for predictions against the protected labels for all datasets. We conduct these analyses to determine whether and where the models were introducing bias and ensure that we only apply our bias mitigation method to aspects of the data that are actually biased. Details of the models can be found in Appendix B.

After performing fairness analysis on each baseline model, we select the protected labels by which our models showed the most bias and test our methods on all datasets. For all fairness analyses in our experiments, we utilize a fairness toolkit developed by IBM - AI Fairness 360 (\cite{bellamy2019ai}).

\subsection{Proposed Method Implementation}

We apply our bias-reducing MTL method by developing MTL models to predict target and protected labels by which the baseline models are biased. We first train the MTL model for each of the combinations for all datasets. We assign different loss weight ratios to the label pairs, prioritizing the prediction label over the protected label to ensure that the model prioritizes the prediction label and makes certain that its uncertainty score is kept lower than the protected label prediction. We train the MTL networks for 100 epochs for all the datasets and save the weights at each epoch, making it a total of 100 weight combinations for each experiment run.

After running the MTL networks and obtaining an ensemble of models, we calculate the uncertainties of these models. We computed the uncertainties by calculating the sample variance of the different forward passes using Equation \ref{eq:uncert}. We build single-task models using the saved model weights to predict the target labels: income for ADULT, IHM for MIMIC-III, and the 4 mood-related labels for SNAPSHOT. With the predictions obtained from these models, we calculate the DIRs. These steps provide us with distributions of 100 performance scores and 100 DIR scores. We implement the non-dominated sorting algorithm introduced in \ref{ssec:nd_sorting} to obtain the sets of Pareto optimal solutions for the experiments for each dataset. This algorithm computes an output containing the models with the best trade-off between the performance score and the fairness score (DIR), and for which there were no better-performing models. Any model can be chosen from these models according to what is being prioritized for the particular problem type and dataset. We plot the performance scores against the DIR scores to get a visual representation of their relationship.  

\subsection{Benchmark Bias Mitigation Approaches}
We compare our proposed method to the following standard and state of the art bias mitigation methods applied in prior work.

(1) Reweighting (\cite{kamiran2012data}) (ADULT, MIMIC-III, and SNAPSHOT): a standard bias mitigation method that has been used extensively in the field and is considered a benchmark technique for bias mitigation.

(2) Adversarial Reweighted Learning (ARL) (\cite{lahoti2020fairness}) (ADULT and MIMIC-III): uses an adversarial setup to reweight features for fairness, which presents a direct method to manipulate feature importance based on their impact on bias, while our approach, using MC Dropout, offers a probabilistic view of feature importance by assessing uncertainty.

(3) FairRF (\cite{zhao2022towards}) (ADULT and MIMIC-III): focuses on mitigating bias through feature manipulation, and provides a direct contrast to our method's use of uncertainty estimation to influence model predictions. This comparison aims to illustrate how managing uncertainty can potentially be more effective than directly modifying feature weights.


Comparing these approaches can highlight the benefits of a probabilistic versus adversarial framework in bias mitigation. See details of each baseline model in Appendix B.

\subsection{Saliency Maps}

To better understand the influence of the input features on our final predictions and also the difference our proposed method makes on how the networks determine what features to prioritize for prediction, we utilize a saliency map technique (\cite{simonyan2013deep}) to visualize the importance of model weights on our input time axis and features. According to the predicted class $c$, for example, the decision-making process of the model can be represented as $S_{c}(I) = w_{c}^{T}I + b_c$, where $I, S_c(I), w, b$ are the model input, output, weights, and bias, respectively. The purpose of this method is to calculate the model weights on the input layer by gradients, for example, $w = \frac{\partial S_c}{\partial I}|_{I_0}$. The calculated $w$ represents the model saliency associated with the input layer. In this study, we fetch the saliency maps for all the test samples and calculate the average saliency map to develop the general intuition of important features and time steps.

We calculate the feature importance for our experiments according to the formulas in this section and plot them in the figures presented in Section \ref{sec:results}.

\section{Results} \label{sec:results}

\subsection{ADULT Dataset} \label{sec:adult_results}

\subsubsection{Fairness Analyses}

The initial analysis before running a model shows that all 3 protected labels were biased by data. For example, for the age label, the DIR score is above 2 for both the training and testing data; for sex, it is approximately 0.36 for both, and for race, it is approximately 0.47 for both. After running the base model, this trend continues, where the model introduces even more bias by age, making the DIR above 3 and below 0.1 for sex, At the same time, for race, it stays similar to the initial analysis with a score of approximately 0.5.

\subsubsection{Methods Implementation}
\begin{figure}[htbp]
\centerline{\includegraphics[width=\textwidth]{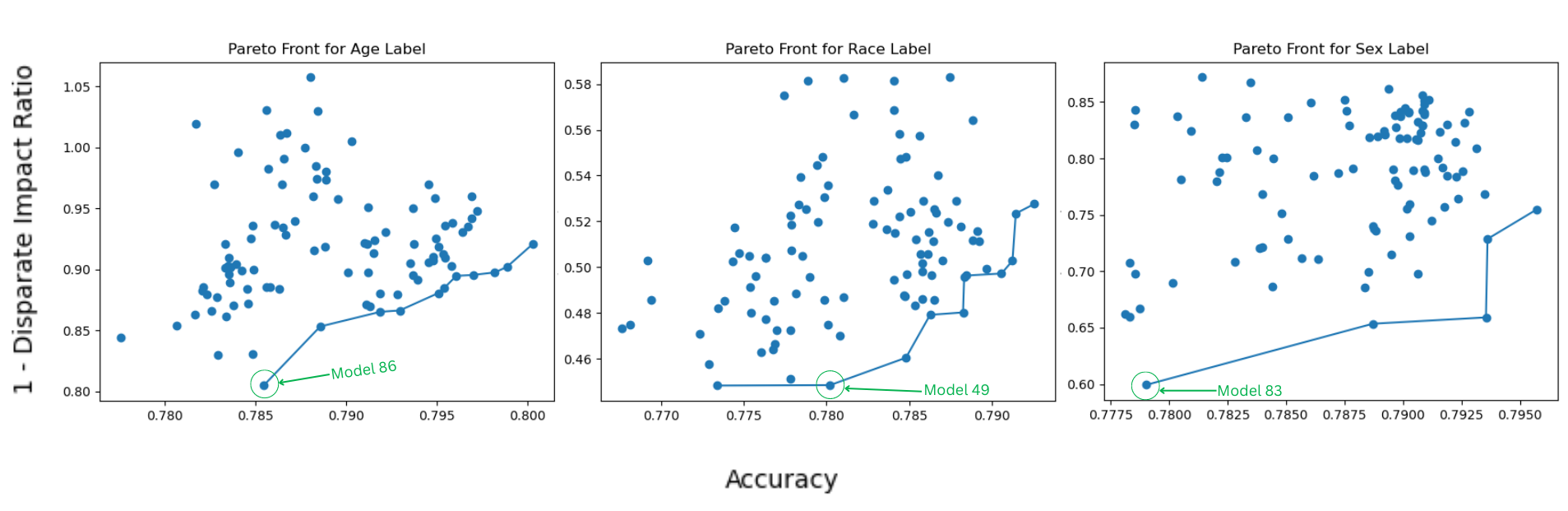}}
\caption{ADULT: Pareto Front Plot for Age, Race, and Sex Labels}
\label{adult_pf_all}
\end{figure}

Since our analyses on the baseline model show bias by all 3 chosen protected labels where their DIRs are beyond the acceptable range of 0.8-1.2, we experiment with our proposed method on all 3 protected labels. After implementing the proposed method and running the non-dominated sorting algorithm on the models we obtained, the Pareto fronts the method produced can be seen in Figures \ref{adult_pf_all}. To give a better, more linear visual representation of our results, we calculate the absolute value of [$1 - $DIR Score] for the models to plot against the accuracy. The lines on the plots represent the Pareto fronts which contain the models that perform best to choose from. For all of the experiments on this dataset, we opt for the models that provide us the best fairness each time due to the insignificant differences in performance (accuracy) between the models (they all fall between 0.77 and 0.8). Therefore, we choose the models enclosed in the green circles to represent our final models.

\begin{figure*}[htbp]
\centerline{\includegraphics[width=\textwidth]{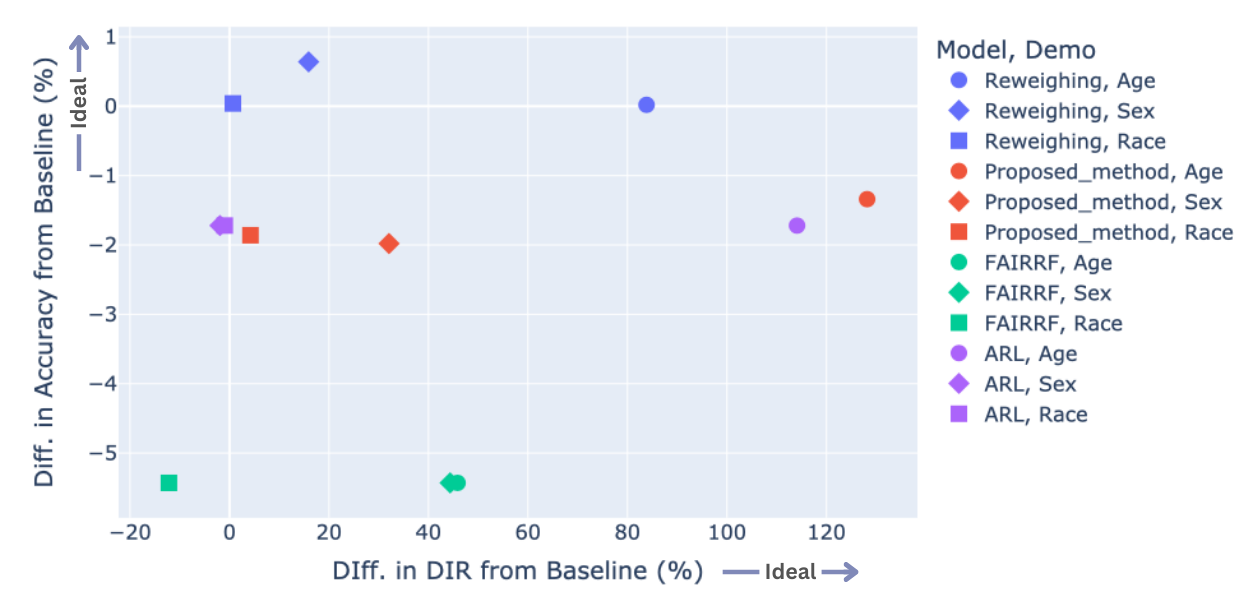}}
\caption{ADULT: DIR and Accuracy Scores of Models}
\label{adult_DI_acc_plot}
\end{figure*}

We compare our final performance results of the proposed method, reweighing, ARL, and FairRF in relation to the baseline performances as shown in Figure \ref{adult_DI_acc_plot}. In this figure, the x-axis represents the percentage improvement in the DIR score from the baseline in relation to the ideal score of 1, and the y-axis shows the percentage differences of the accuracy scores from the baseline accuracy. The reweighing technique improved the fairness scores, with the least impact on the accuracy of the models. For the protected age label, reweighing improved the DIR score by 84\%, for sex by 15.9\%, and for race by 00.67\% which we deemed not significant. FairRF improved fairness for age and sex by approximately 45\%, better than all other methods, but at a higher cost of performance. For the race label, FairRF diminished both fairness and accuracy of the model.

The chosen model from our proposed method performed better than reweighing in terms of fairness, with an improvement in the DIR score of the age label by 128\%, sex by 32\%, and 4.2\% for the race label. However, this improvement in fairness came at a slight performance cost, where the proposed method caused an average decrease of 1.7\% in accuracy for all 3 labels. The reweighing method had an even lower performance cost, improving the accuracy score for the sex label. Our implementation of ARL did not improve fairness of the sex and race labels, but significantly improved the DIR score for the age label by 114\%, which is most comparable to the proposed method. This method also produced similar accuracy scores to the proposed method.


\subsubsection{Explainability}
\begin{figure*}[htbp]
\centerline{\includegraphics[width=\textwidth]{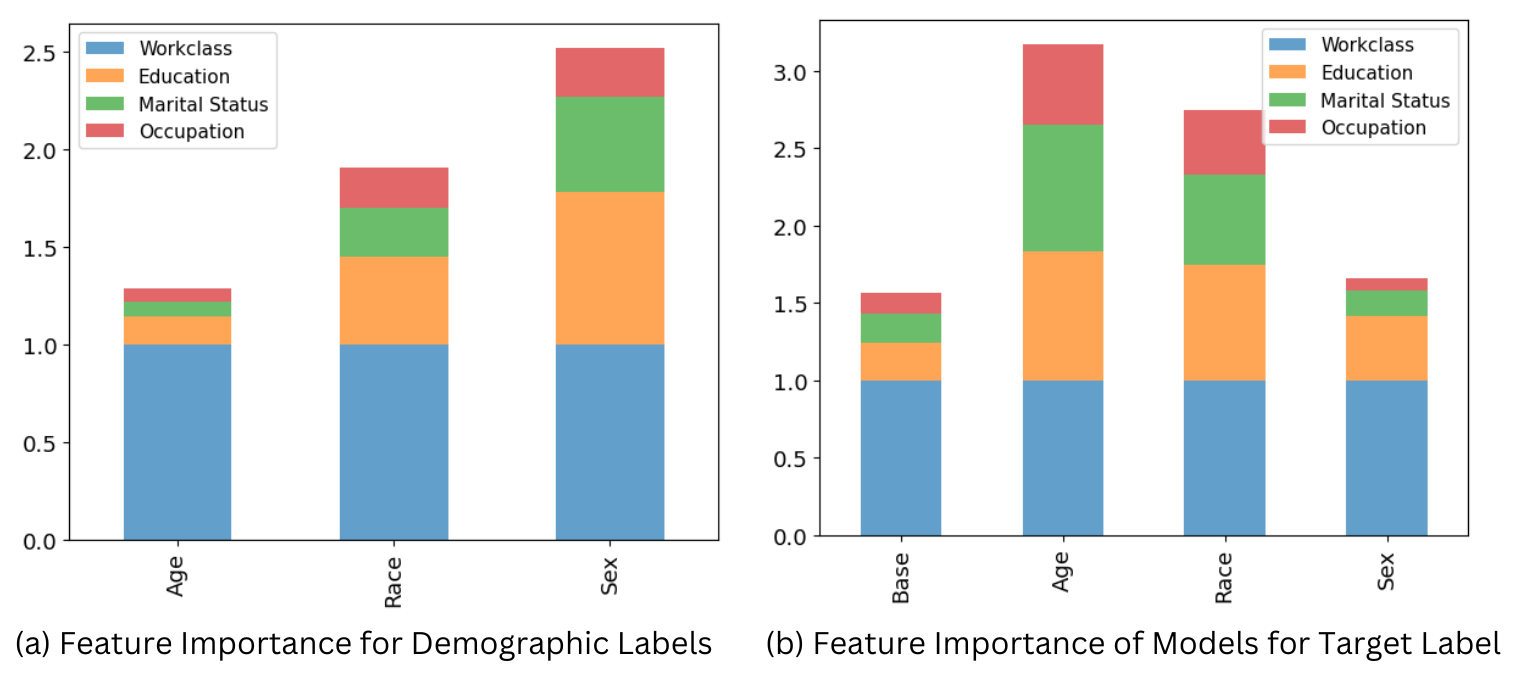}}
\caption{ADULT: Feature Importance Comparison Between Protected and Target Labels}
\label{adult_feat_imp_comp}
\end{figure*}

After calculating the feature importance for the baseline model and the models produced by our proposed method for age, race, and sex labels, the scores we obtained show us that our method redistributed weights to features less relevant to the protected label. This change is demonstrated in Figure \ref{adult_feat_imp_comp}. This figure visualizes the feature importance for the demographic labels as bar charts, which were chosen over saliency maps due to their clearer representation in scenarios without temporal data dependencies, making them easier to understand and interpret. The figure visualizes the feature importance for the demographic labels, and b represents the feature importance for the target label from the implementation of the base mode, and those of our proposed method for each of the protected labels. These figures show that the workclass feature holds the most weight across the board for all labels and remains equally important after the proposed method implementation. On the other hand, the relationship feature holds no weight in any of the prediction tasks across all experiments. The changes in importance are hence demonstrated more on the 3 remaining features: education, marital status, and occupation. Furthermore, the relative importance of these features changes across the experiments according to the demographic label we tested our proposed method on. For example, for the age label, the least important features are occupation, marital status, and education in that order, and the proposed method implemented with age shifted weight importance to these features. This trend is observed for the other labels of race and sex. 

\subsection{MIMIC-III Dataset}

\subsubsection{Fairness Analyses}
The baseline model for IHM prediction using the MIMIC-III dataset results in Accuracy = 0.866, AUROC = 0.603, and AUPRC = 0.181. Our analysis of the MIMIC-III data shows that the baseline model is only biased by marital status out of the 4 protected labels we analyzed, with a DIR score of 1.3. Regarding age, gender, and insurance labels, the baseline model shows unsubstantial bias where the DIRs calculated for each fell within the acceptable range of 0.8-1.2, as shown in Table \ref{mimic_fairness_scores}. The differences in FN scores are also significantly low across all the protected labels. 

\begin{singlespace}

\begin{table}[htbp]
\centering
\caption{MIMIC: Fairness Scores for Baseline Model}
\begin{tabular}{|l|l|l|}
\hline
\textbf{Label} & \textbf{DI Ratio Score} & \textbf{Diff in FN Scores} \\ \hline
Age & 1.008 & -0.006 \\ \hline
Marital status & \cellcolor[gray]{0.8}1.308 & 0.005 \\ \hline
Gender & 1.182 & -5.645${\rm e}^{-05}$\\ \hline
Insurance & 1.021 & -0.004  \\ \hline
\end{tabular}
\label{mimic_fairness_scores}
\end{table}
\end{singlespace}

\subsubsection{Methods Implementation}
\begin{figure*}[htbp]
\centerline{\includegraphics[scale=0.4]{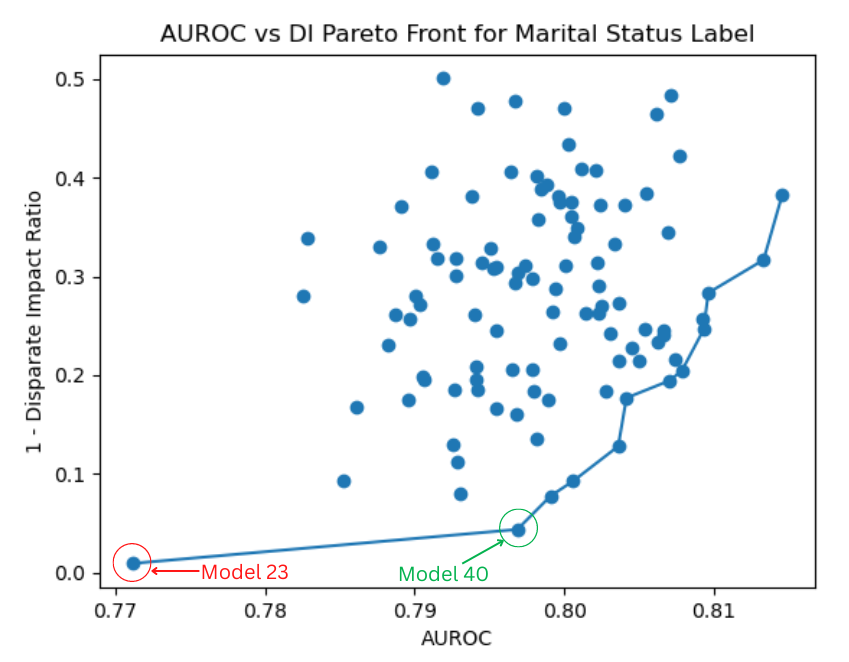}}
\caption{MIMIC-III: Pareto Front Plot for AUROC vs DI Ratio Score}
\label{mimic_pf_auroc}
\end{figure*}

Since our analyses show bias by marital status, we experiment with our proposed method using our target label of IHM and the protected label, marital status. After implementing the proposed method and running the non-dominated sorting algorithm on the models we obtained, the Pareto fronts the method produced can be seen in Figure \ref{mimic_pf_auroc} for DIR plotted against the chosen metric AUROC. To provide a better, more linear visual representation of our results, we calculate the absolute value of [$1 - $DIR Score] for the models to plot against the accuracy as we did in \ref{sec:adult_results}. The lines on the plots represent the Pareto fronts which contain the models that performed best to choose from for each of the performance scores. Our proposed method improves the 3 performance scores for IHM prediction across the 4 protected labels. The general accuracy for the models on the Pareto front shows an average improvement of 1.74\% from the baseline model. The AUROC score improves by an average of 20\%, and the AUPRC score by 19.7\%. In this experiment, we focus on prioritizing the AUROC score, as mentioned in Section \ref{sec:base_models} because AUROC is prioritized for this dataset in previous studies (\cite{harutyunyan2019multitask}), and the accuracy variability for the models in the Pareto front is low (approximately 0.65\%). We select our final model from the AUROC vs. DI score Pareto plot (Figure \ref{mimic_pf_auroc}). From this plot, the 2 circled models represent the optimal solutions from our method. We choose model 40 as our final solution since it provided an almost equal improvement in DIR as model 23, with a more significant improvement in AUROC, AUPRC, and accuracy. 

Implementing the reweighing bias mitigation technique shows similar prediction performance when compared to our proposed method (as seen in Table \ref{mimic_pareto_scores}), improving the AUROC, AUPRC, and accuracy slightly less than the proposed method when compared to the baseline. Reweighing improves the DIR score of the baseline model by 12\% but does not perform as well as our proposed method. Regarding the difference in FN scores, the reweighing technique performs worse than the baseline. Implementation of ARL and FairRF performed comparably to the proposed method in terms of fairness considering DIR scores, but they diminished performance at significantly higher rates. There was an average drop of 7.9\% in AUROC between the 2 methods.

\begin{singlespace}
\begin{table}[htbp]
\centering
\caption{MIMIC: Performance and Fairness of Models from Proposed Method, Reweighing Method, and Baseline Model}
\begin{tabular}{|l|l|l|l|l|l|l|}
\hline
\textbf{Method} & \textbf{Acc} & \textbf{AUROC} & \textbf{AUPRC} & \textbf{DIR} & \textbf{Diff in FN} \\ \hline

\textbf{Base Model} & 0.886 & 0.603 & 0.181 & 1.308 & 0.005 \\ \hline

\textbf{Reweighing} & 0.883 & 0.777 & 0.332 & 1.188 & -0.05 \\ \hline

\textbf{ARL} & 0.719 & 0.528 & 0.136 & 0.954 & 0.064 \\ \hline

\textbf{FairRF} & 0.669 & 0.520 & 0.118 & 0.982 & -0.035 \\ \hline

\cellcolor[gray]{0.8} \textbf{Proposed Method} & \cellcolor[gray]{0.8} 0.883 & \cellcolor[gray]{0.8} 0.797 & \cellcolor[gray]{0.8} 0.367 & \cellcolor[gray]{0.8} 1.043 & \cellcolor[gray]{0.8} 0.005 \\ \hline

\end{tabular}
\label{mimic_pareto_scores}
\end{table}
\end{singlespace}

\subsubsection{Explainability}
The saliency maps from experiments on the MIMIC-III dataset can be found in Appendix C. The deeper red color in these figures represents features that hold more weight and, therefore, are more critical for target prediction. Evaluating the map for the MIMIC-III baseline model gives us an insight into some of the features that carried the most weight for IHM prediction. For example, the Glascow coma scale (GCS) motor response for Abnormal flexion and extension appears to be highly correlated with IHM prediction. GCS is a tool used to assess a patient's level of consciousness after a brain injury (\cite{jain2018glasgow}). The GCS motor response score evaluates the patient's ability to obey commands or move in response to a stimulus. Numerous studies have examined the relationship between the GCS motor response score and in-hospital mortality (\cite{yumoto2019association, sadaka2012four, settervall2011hospital}). Generally, lower motor response scores are associated with a higher risk of mortality which explains the relevance of these features in our IHM prediction model.

Two other features that are shown to be highly correlated with the baseline IHM prediction are Fraction inspired oxygen (FiO2), which is a measure of the concentration of oxygen that a patient is receiving, usually through supplemental oxygen therapy (\cite{fuentes2020fraction}), and pH, which is a measure of the acidity or alkalinity of a patient's blood (\cite{arias2016neonatal}). Evidence suggests that higher FiO2 levels may be associated with an increased risk of in-hospital mortality (\cite{de2008association}), where higher FiO2 levels were associated with an increased risk of mortality in both adult and pediatric patients. There is also some evidence from prior studies that indicate an association between abnormal pH levels and an increased risk of in-hospital mortality (\cite{park2020normothermia}), where both low and high pH levels were associated with increased mortality in critically ill patients.

\subsection{SNAPSHOT Dataset}

\subsubsection{Fairness Analyses}
\begin{figure*}[htbp]
\centerline{\includegraphics[width=\textwidth]{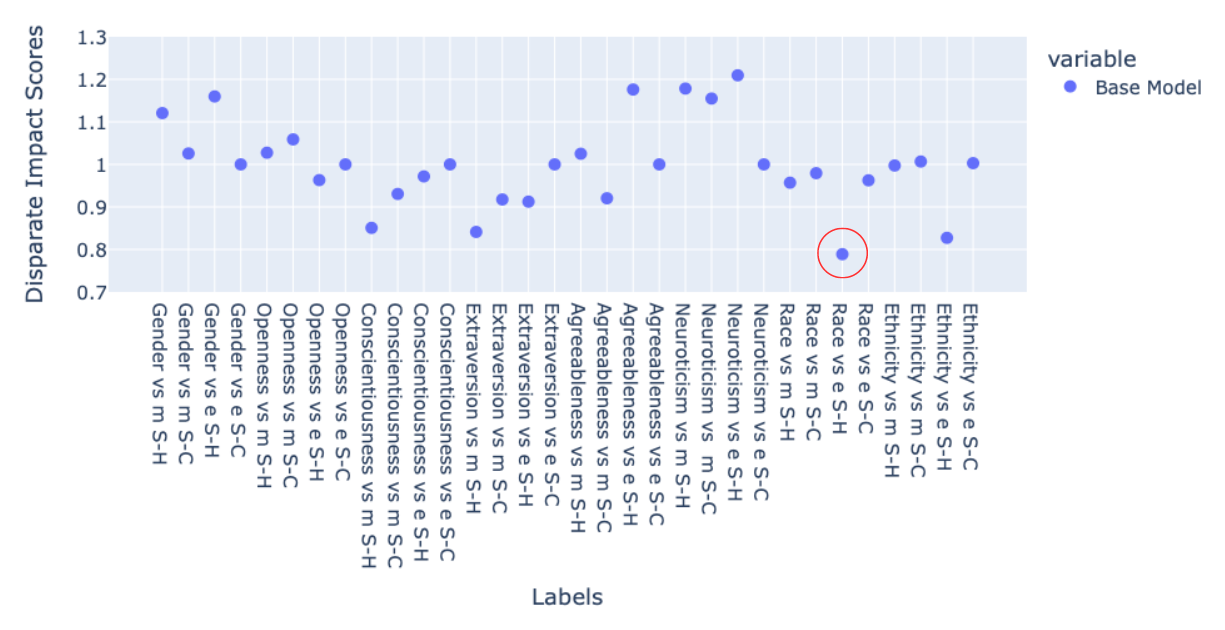}}
\caption{SNAPSHOT: Scatter Plot of DIR Scores from Baseline Model}
\label{snap_initial_DI}
\end{figure*}

After performing the initial set of 32 fairness analyses on the SNAPSHOT data, our results show that the data itself does not contain substantial bias by any of the 8 protected labels we tested by. The scatter plot in Figure \ref{snap_initial_DI} shows that all scores fall within the acceptable range of 1.2 and 0.8 for the DIR score, except for the experiment with Race as a protected label and evening-sad-happy as the target label (specified by the red circle on the figure). For the 4 target labels, morning-sad-happy, morning-stressed-calm, evening-sad-happy, and evening-stressed calm, the F1 performance scores fell from approximately 0.7 to 0.8.

\subsubsection{Methods Implementation}
\begin{figure*}[htbp]
\centerline{\includegraphics[scale=0.4]{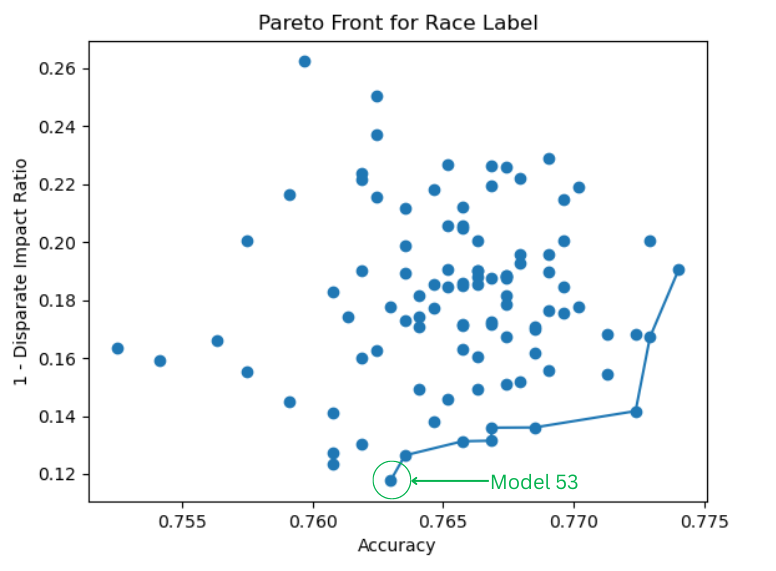}}
\caption{SNAPSHOT: Pareto Front Plot for Race vs Evening-sad-happy Label}
\label{snap_pf_race}
\end{figure*}

We run our proposed method on Race as a protected label and evening-sad-happy as the target label and obtain accuracy scores in the range of 0.76 to 0.77 and DIR scores between 0.80 and 0.88 for the model ensemble produced. We choose the model with the best DIR score of 0.882 as shown in Figure \ref{snap_pf_race}. This is an improvement of approximately 0.1 in fairness from the baseline model with a DIR score of 0.78. In terms of performance, this method did not affect the F1 score when compared to the baseline model. Both have F1 scores of approximately 0.76. Using the reweighing technique, we obtain a perfect DIR score of 1, but that comes at a high cost of performance, with the F1 score dropping to 0.39, a loss of 37\% in performance.

\subsubsection{Data Manipulation on SNAPSHOT}

After analyzing the SNAPSHOT baseline model for bias
and our results showing that the data is only biased by the Race label for evening-sad-happy label prediction; we perform some data manipulation to create a more significant imbalance for the gender label in order to determine if this would decrease the fairness performance and to test our proposed method on it. Details on this process can be found in Appendix B.

After running the base model for all 4 target labels on this newly manipulated data and analyzing fairness for all 3 experiment settings, we obtain similar performance scores. The fairness scores for experiments S-1 and S-2 are similar, with only the DIR score of the target label of evening-sad-happy occurring outside of the acceptable bounds of 0.8-1.2. In contrast, the fairness scores for experiment S-3 did not change significantly from the original SNAPSHOT experiment with no data manipulation. The F1 and fairness metric scores for experiment S-2 are listed in Table \ref{snap_altered_results}. As seen in the table, the DIR score of the target label of evening-sad-happy (the gray highlighted cell) falls outside the acceptable bounds of 0.8 and 1.2. All other fairness scores fall within these acceptable bounds.

\begin{singlespace}
\begin{table}[htbp]
\centering
\caption{SNAPSHOT: F1 and Fairness Scores for Altered Data Experiments}
\begin{tabular}{|l|l|l|l|l|}
\hline
\textbf{Label} & \textbf{F1 Score} & \textbf{DI Ratio Score} & \textbf{Diff in FN} & \textbf{Diff in FP} \\ \hline
Morning-Sad-Happy & 0.749 & 1.153 & -0.041 & 0.0979 \\  \hline
Morning-Stressed-Calm & 0.803 & 1.045 & -0.009 & 0.073 \\ \hline
Evening-Sad-Happy & 0.798 & \cellcolor[gray]{0.8}1.291 & -0.123 & 0.088 \\ \hline
Evening-Stressed-Calm & 0.788 & 1.0536 & -0.0281 & 0.048 \\ \hline
\end{tabular}
\label{snap_altered_results}
\end{table}
\end{singlespace}

\begin{figure*}[htbp]
\centerline{\includegraphics[scale=0.4]{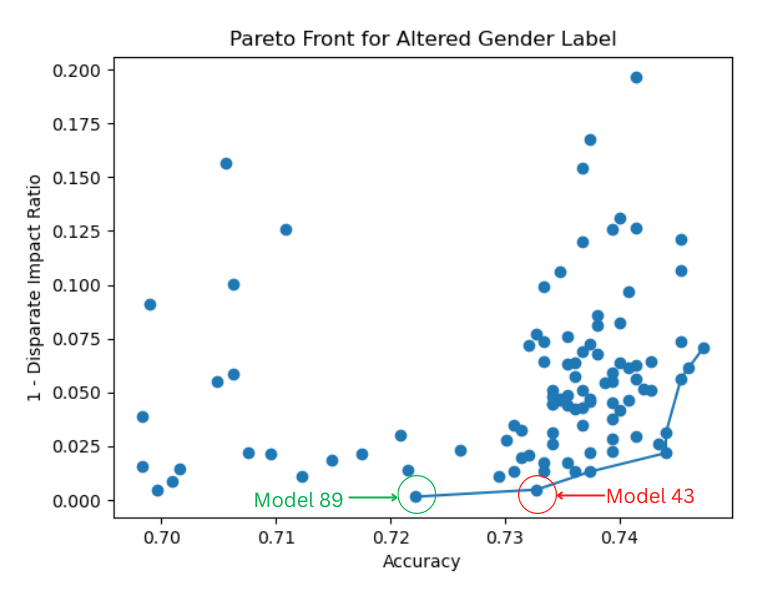}}
\caption{SNAPSHOT: Pareto Front Plot for Gender vs Evening-sad-happy Label with Altered Data}
\label{snap_pf_gender_altered}
\end{figure*}

After running our proposed bias mitigation algorithm on the label combination of evening-sad-happy and gender on the manipulated data, we obtain an ensemble of models whose performance and DIR scores are plotted in Figure \ref{snap_pf_gender_altered}. On average, the models have an F1 score of 0.74 which is a 5.8\% drop in performance from the baseline evening-sad-happy model. The average DIR score of the 9 models on the Pareto front is 1.034, which is well within the acceptable bounds of 0.8-1.2 and an overall improvement of 0.257 from the base DI ratio score of 1.291. We chose models 89 and 43 (green and red circles in Figure \ref{snap_pf_gender_altered}) as potential final models. Model 89 has the best DI ratio score of 1.001, which is almost the ideal score of 1, but it comes with a diminished F1 score of 0.722, a 7.6\% drop in performance. It differs in FP and FN rates scores of -0.082 and -0.084, respectively. Model 43 has a DI ratio score of 0.995, with slightly better performance (0.733) than model 89. 

\subsubsection{Explainability}

The saliency maps from experiments on the SNAPSHOT dataset can be found in Appendix C. The deeper red color in these figures represents features that hold more weight and are, therefore, more important for target prediction across the 7 timesteps. From the saliency maps for the baseline method on SNAPSHOT, a few features show the highest weights in the prediction task, at the top of this list of features being no\_sleep\_24, phys\_3H-10H-medStepsWeightedPeaks, SMS\_0H-3H\_total\_num. Comparing these maps to those from the model produced by our proposed method on SNAPSHOT, it can be noted that our proposed method distributed the weights more evenly among all the features in general, explicitly moving some of the importance away from the more highly-weighted features in the baseline model prediction.

\section{Discussion}

\subsection{Fairness and Performance}

Facing the recurring challenges of bias and fairness in machine learning, our study introduces a bias mitigation technique that allows for tuning the trade-off between fairness and performance. Our method demonstrates enhanced fairness across three datasets—ADULT, MIMIC-III, and SNAPSHOT with minimal impact on performance and notable improvements for MIMIC-III. When compared to the standard reweighing bias mitigation technique, our approach shows significant improvements in fairness for the ADULT and MIMIC-III datasets and a slight improvement for SNAPSHOT, all without sacrificing performance. In contrast, reweighing reduces the F1 score by 37\%. Overall, our method outperforms other state-of-the-art methods implemented on ADULT and MIMIC-III datasets in boosting fairness while maintaining performance.

From our experiments in predicting income with the ADULT dataset, our analyses show us that the baseline model prediction of income was biased based on age, sex, and race, which corresponds with analyses from prior works (\cite{le2022survey, lahoti2020fairness, friedler2019comparative, zafar2017fairness}). Implementing our proposed method provides a distribution of models that improves the DIR scores to within the ideal threshold of 0.8 and 1.2. This method performs better than the reweighing technique in terms of fairness, leading to an average of 1.7\% decrease in accuracy for all three protected labels.

The LSTM model predicting in-hospital mortality (IHM) using the MIMIC-III dataset exhibits bias by marital status but not by age, gender, or insurance type. This finding aligns partially with the fairness analysis by \cite{roosli2022peeking}, which noted no gender bias. Unlike their study, which did not consider marital status, we include it based on evidence suggesting its significant health impact. Studies have shown that unmarried individuals face higher rates of adverse cardiovascular events and mortality compared to married ones (\cite{dhindsa2020marital, johnson2000marital}).

Upon identifying marital status bias, we apply our proposed bias mitigation technique, enhancing fairness to nearly optimal levels with a DIR of 1.04 and substantially equalizing FN rates across groups. Additionally, our method boosts model performance, increasing accuracy by 1.7\%, AUROC by 19.4\%, and AUPRC by 18.6\%. This surpasses the performance improvements of the reweighing technique. Our findings support research suggesting that does not always lead to a drop in utility. Instead, it could lead to enhanced performance under specific circumstances (\cite{deho2022existing}).

The ARL method shows promising results in improving fairness based on the age label for the ADULT dataset, but it falls short in addressing fairness concerns related to sex and race labels. This discrepancy could be linked to challenges similar to those observed with datasets like COMPAS, where ARL struggles to effectively address race-related fairness issues in the original publication. This indicates potential gaps in the ARL approach where race or other protected attributes are not computationally identifiable. Notably, the publication does not provide a detailed explanation for the method's under-performance on certain datasets and attributes.

The FairRF method, developed by Zhao et al., introduces an approach that utilizes related features to develop fair classifiers without direct reliance on sensitive attributes. This method significantly enhances fairness for age and sex labels on the ADULT dataset by approximately 45\%, surpassing other methods but at a considerable performance cost. However, for the ADULT race label, it reduces both fairness and accuracy, while it improves fairness for marital status on the MIMIC-III dataset, also at a significant performance cost. A critical limitation of the FairRF method is its heavy dependence on the selection and quality of related features. If the set of related features is incomplete or comprises noisy data, the effectiveness of the FairRFmethod may be significantly reduced. The success of this approach is highly dependent on how accurately the related features can represent the underlying sensitive attributes without introducing new biases.

For SNAPSHOT, we analyze baseline models that predict 4 self-reported mood-related labels to determine if they are biased by 3 demographic labels, namely gender, race, and ethnicity, and 5 personality-based labels, namely openness, conscientiousness, extraversion, agreeableness, and neuroticism. These analyses determine that the DIR scores from all our experiments are within the chosen acceptable range of 0.8 to 1.2, except for the model predicting evening-happy-sad when analyzed by race. Implementing our proposed method improves the DIR score to 0.88, an improvement of 0.1 from that of the baseline method. In terms of performance, our proposed method has little effect in this case. Implementing the reweighing technique provides a perfect DIR score of 1 but a considerable performance cost of 37\%.

As our initial analyses on the SNAPSHOT dataset baseline models do not show substantial bias by the 8 protected labels, we manipulate the data by removing samples from 75 participants identified as female from the training data to determine the performance of our proposed method further. After running the baseline model on this altered data and analyzing, it shows bias by gender for the evening-sad-happy label with a DIR score of 1.29, slightly above our chosen acceptable range. Running our proposed technique results in an average, almost ideal DIR score of 1.034, an average 5.8\% drop in performance.

\subsection{Improving Model Explainability}

In this work, we propose a method that not only enhances fairness in machine learning models but also aids practitioners across various sectors in better understanding the decision-making processes of their models. Recognizing that simply reducing model bias may not fully address fairness concerns, especially in complex fields like predictive healthcare which can be complex, unintuitive, and often hard to explain, our method focuses on improving model explainability to increase the utility of outputs and achieve better outcomes (\cite{yang2022explainable}). By demonstrating which features most influence model predictions and adjusting feature weight away from protected labels, our approach promotes transparency and mitigates disparate treatment of unprivileged groups.

Our experiments demonstrate these concepts effectively. For instance, using the ADULT dataset, Figure \ref{adult_feat_imp_comp} indicates our method reduces reliance on features closely associated with protected labels, thereby minimizing their impact on predictions. Similarly, saliency maps for the MIMIC-III and SNAPSHOT datasets, detailed in Appendix C, show our method’s ability to redistribute feature importance, ensuring a more balanced consideration of inputs unrelated to protected attributes. This alignment with our hypothesis confirms that de-emphasizing features correlated with protected labels diminishes bias without significantly compromising model performance.

Additionally, the explainable characteristic of our work also eliminates the need for researchers to initially identify the undesired correlation between the target label and the protected label to be able to tackle bias in the model using general existing bias mitigation methods (\cite{wang2020towards}).

\subsection{Performance-Fairness Trade-off}

Our method employs non-dominated sorting and the concept of Pareto optimality from multi-objective optimization to enhance the performance-fairness trade-off. This approach has allowed us to identify a set of Pareto-optimal solutions, presenting a range of models from which proprietors can select based on their specific needs. This method offers flexibility, enabling users to prioritize either fairness or performance according to the application or data type. In our experiments, we leverage this feature to select models with the highest fairness scores, as the variability in performance is consistently low across all tested scenarios.

\subsection{Limitations}

One limitation of this research is the restriction to binary outcomes imposed by the tools and techniques utilized for bias mitigation and the outcome labels. Another limitation is the general lack of an agreed-upon definition of fairness in the field (\cite{park2020bias}). In the area of Machine Learning, most attempts at proposing new fairness constraints for algorithms have come from the West, and most of the works use the same datasets and problems to show how their constraints perform, which means that there is still no explicit agreement on which constraints are the most appropriate for which problems (\cite{mehrabi2021survey}). Even though fairness is an incredibly desirable quality in society, it can be surprisingly difficult to achieve in practice. With these challenges in mind, there needs to be more standardization regarding fairness definitions and a more rigorous evaluation of what definitions should be applied to different data and problem types. In the future, it will be interesting to determine how popularly-used definitions of fairness translate to models built for data from different domains and how effective they are for each of these models in real-world scenarios.  

Working with deep learning models, we have also encountered limitations associated with the computational complexity of using MC Dropout. Using this method along with the aspect of saving and loading model weights results in a higher than normal computational cost and time. 

\section{Conclusion}

Our proposed method leverages feature weights for bias mitigation and has been successfully applied to data from three different jurisdictions—enhancing fairness across the ADULT, MIMIC-III, and SNAPSHOT datasets. It achieves improvements in fairness at a rate comparable to or greater than that of Reweighing, a standard bias mitigation technique, and other existing methods, without significant sacrifices in performance. Additionally, our approach offers flexibility in managing the fairness-performance trade-off, utilizing principles from multi-objective optimization to allow users to prioritize their desired objectives.

Furthermore, our method enhances model transparency through the use of saliency maps, which detail shifts in feature importance for prediction tasks. This transparency is crucial for building trust among professionals in various fields, such as healthcare, encouraging them to integrate these predictive models into their daily operations (\cite{yang2022explainable}).

\vspace*{0.05in}

\vspace*{0.5in}

\bibliographystyle{unsrtnat}
\bibliography{references}  

\clearpage 

\appendix
\section*{Appendix A. Data Description}

In this appendix we provide further description of the ADULT, MIMIC-III and SNAPSHOT datasets and our data processing procedures.

\subsection{ADULT Data} 

The ADULT dataset was drawn from the 1994 United States Census Bureau data and involves using personal details such as education level to predict whether an individual will earn more or less than \$50,000 per year (\cite{kohavi1996scaling}). This dataset has been extensively utilized in machine learning bias and fairness research (\cite{zafar2017fairness, friedler2019comparative, kearns2019empirical, du2021fairness, fabris2022algorithmic}). The extracted features from the dataset used in this work are summarized in Table \ref{tab:adult}.

\begin{singlespace}
\begin{table}[htbp]
\centering
\caption{ADULT: List of extracted features}

\begin{tabular}{|c|c|}
\hline
Feature                 & Classes                                                                                                                      \\ \hline
Age &   Continuous                                                               \\ \hline



Education   & \begin{tabular}[c]{@{}c@{}} Preschool, 1st-4th, 5th-6th, 7th-8th, 9th, 10th, 11th, 12th, \\ HS-grad, Some-college, Assoc-voc, Assoc-acdm, Prof-school, Bachelors, \\ Masters, Doctorate \end{tabular}          \\ \hline




Marital Status     & \begin{tabular}[c]{@{}c@{}} Married-civ-spouse, Divorced, Never-married, Separated, Widowed, \\ Married-spouse-absent, Married-AF-spouse  \end{tabular}
    \\ \hline


Occupation  & \begin{tabular}[c]{@{}c@{}} Tech-support, Craft-repair, Other-service, Sales, Exec-managerial, \\ Prof-specialty, Handlers-cleaners, Machine-op-inspct, Adm-clerical, \\ Farming-fishing, Transport-moving, Priv-house-serv, Protective-serv, \\ Armed-Forces \end{tabular}
    \\ \hline

Race    & White, Asian-Pac-Islander, Amer-Indian-Eskimo, Other, Black
    \\ \hline
    
Relationship    &  Wife, Own-child, Husband, Not-in-family, Other-relative, Unmarried
    \\ \hline

Sex & Female, Male
    \\ \hline

Workclass      & \begin{tabular}[c]{@{}c@{}} Private, Self-emp-not-inc, Self-emp-inc, Federal-gov, State-gov, \\ Without-pay, Never-worked \end{tabular} 
    \\ \hline

\end{tabular}
\label{tab:adult}
\end{table}
\end{singlespace}

This dataset contains 14 categorical and continuous integer attributes from which we dropped columns with noisy data (over 100 missing values). We were left with 8 features: Age, Workclass, Education, Marital status, Occupation, Relationship, Race, and Sex. 

The prediction task for this dataset is a binary income/salary prediction, where a label of 1 represents participants who earn $>$ 50k and 0 for those who earn $<=$ 50k in a year. We converted all the string features to integers, and we used 45222 of 48842 instances after dropping a total of 3620 samples with noisy data. We utilized age, sex (gender), and race as the protected labels in our experiments. 

For fairness analysis, we binarized the 3 protected labels to 0 as unprivileged and 1 as privileged classes. For the age label, 12\% of the female class earned more than 50k in salary, while for the male class, 32\% earned more than 50k. 36\% of participants above the age of 40 earned more than 50k, while 17\% of those below the age of 40 earned more than 50k. As for the race label, less than 13\% of the participants who identified as Black, Amer-Indian-Eskimo, and "Other" earned more than 50k, while about 30\% of those who identified as White or Asian-Pacific -Islander earned more than 50k. We chose the class corresponding to higher income/salary as the privileged class for each protected label. These privileged classes are also the majority classes in the dataset. For the age label, since the difference in the number of participants between classes is larger than the difference in average salary earned, we chose the class with more participants (less than or equal to 40) as the privileged class (1), while the class with fewer participants (over 40) as the unprivileged class (0). For race, Black(0), Amer-Indian-Eskimo(4), and other(2) became 0, and  White(3) and Asian-Pac-Islander(1) became 1, and for sex, male became 1 and female became 0. We excluded the protected labels as features to feed into our prediction model, and split the data into 75\% training and 25\% testing sets.

\subsection{MIMIC-III Data Description} \label{MIMIC_supplemental}

The MIMIC-III (‘Medical Information Mart for Intensive Care’) data set is a large, freely-available database that contains de-identified health-related data associated with over 40,000 patients who stayed in critical care units of the Beth Israel Deaconess Medical Center. It contains detailed information regarding the care of patients, which includes vital signs, medications, laboratory measurements, observations and notes charted by care providers, fluid balance, procedure codes, diagnostic codes, imaging reports, hospital length of stay, survival data, patient demographics and in-hospital mortality (IHM), laboratory test results, billing information, etc. The complete data description can be found in the publication (\cite{johnson2016mimic}).

\begin{figure*}[htbp]
\centerline{\includegraphics[width=\textwidth]{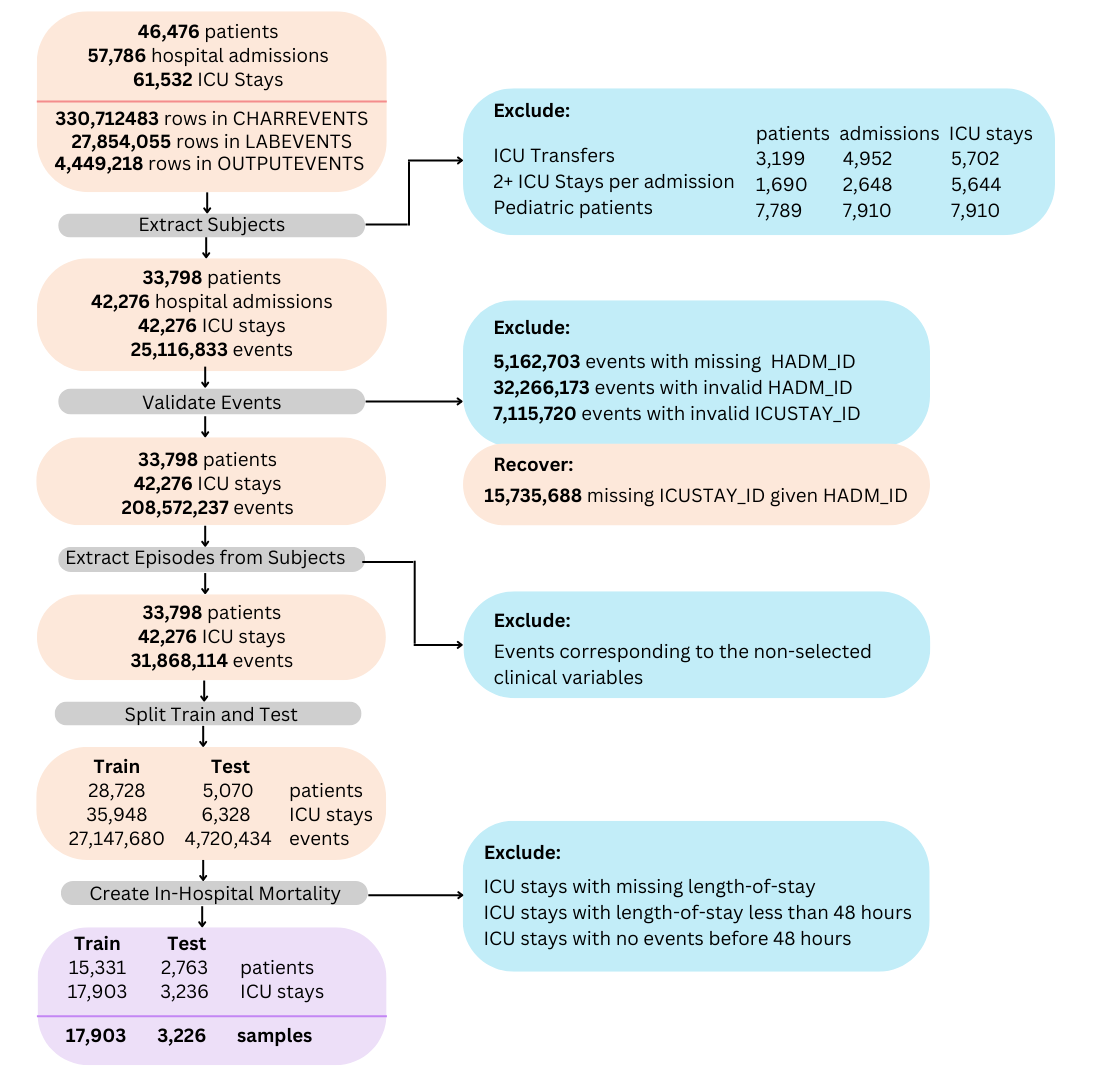}}
\caption{MIMIC-III: Data Preparation Workflow}
\label{fig-mimic-data}
\end{figure*}

The risk of mortality is often formulated as binary classification using observations recorded from a limited window of time following admission (\cite{harutyunyan2019multitask}), and for our experiments on MIMIC-III, we predicted IHM as a binary label, so the target label indicates whether the patient died before hospital discharge. We processed and cleaned the data according to the method in \cite{harutyunyan2019multitask}, where there are two major steps in the data processing pipeline. First, the root cohort is extracted based on the following exclusion criteria: Hospital admissions with multiple ICU stays or ICU transfers are excluded to reduce any possible ambiguity between outcomes, and ICU stays. Patients younger than 18 are excluded as well. Lastly, event entries are only retained if they can be assigned to a hospital and ICU admission and are part of the list of 17 physiological variables used for modeling, namely: Capillary refill rate, Diastolic blood pressure, Fraction inspired oxygen, Glascow coma scale, eye opening, motor response, verbal response, glucose, heart rate, height, mean arterial pressure, Oxygen saturation, respiratory rate, Systolic blood pressure, temperature, weight, and pH (\cite{roosli2022peeking}). From the root cohort, the IHM cohort was extracted by further excluding all ICU stays for which "length of stay" is unknown or less than 48 hours or for which there are no observations in the first 48 hours. The complete process of patient filtering is described in Figure \ref{fig-mimic-data} (\cite{harutyunyan2019multitask}). This provided final training and test sets of 17,903 and 3,236 ICU stays, respectively. Similar to \cite{harutyunyan2019multitask}, We determined IHM by comparing patient date of death with hospital admission and discharge times. The resulting mortality rate is 13.23\% (2,797 of 21,139 ICU stays) (\cite{harutyunyan2019multitask}).

The MIMIC-III dataset has some missing data in some instances. For these, we imputed the missing values using the most recent measurement value if it exists and a pre-specified “normal” value otherwise. These pre-specified values are provided by the authors in \cite{harutyunyan2019multitask}. For categorical values, we encoded them using a one-hot vector and standardized the numeric inputs by subtracting the mean and dividing by the standard deviation. We calculated the statistics per variable after the imputation of missing values. After inputting the missing values and standardizing, we obtained  17 pairs of time series for each ICU stay. We have a binary target label for IHM for each stay, determining whether or not a patient dies in the hospital.  

From this dataset, the demographic data we utilized in our experiments include age, gender, marital status, and insurance, referred to as the protected labels in this paper. Table \ref{mimic_demo_dist} shows the distribution of patients, ICU stays, and IHM rates in our data based on the different demographic groups. The age distribution is skewed towards older patients, with over 80\% above 50 years. As for ethnicity, two-thirds of the patients are non-Hispanic White. In terms of health insurance, According to \cite{roosli2022peeking}, health insurance type is included as a socioeconomic proxy, and this led to us including it in our experiments. In general, three major insurance types are available in the United States: The public programs Medicare and Medicaid, and private insurance providers. Medicare is a federal social insurance program, and anyone over 65 years or with certain disabilities qualifies for it, whereas Medicaid provides health insurance to very low-income children and their families.  Insurance data was therefore mapped to the four distinct categories Medicare, Medicaid, private, and other insurance. In this dataset, roughly a third of the patients are covered by private insurance, and more than 50\% are enrolled in Medicare. The overall IHM rate in the MIMIC-III dataset is 13.23\%, and IHM increases with age. As for gender, female patients appear to have a slightly higher IHM risk, and there is also a large range of variability for ethnic and socioeconomic groups. 

\begin{singlespace}
\begin{table}[htbp]
\centering
\caption{MIMIC-III: Demographic Distributions}
\begin{tabular}{|l|l|l|}
\hline
& \textbf{Patients n(\%)}  & \textbf{IHM rate (\%)} \\ \hline
\textbf{Totals} & 18,094 & 13.23 \\ \hline

\textbf{Age} \\ \hline
0-17 & 0 (0.0)  & 0 \\ \hline
18-29 & 782 (4.3)  & 5.6 \\ \hline
30-49 & 2,680 (14.8) & 9.3 \\ \hline
50-69 & 6,636 (36.7) & 11.1 \\ \hline
70-89 & 7,043 (38.9) & 16.5 \\ \hline
90+ & 953 (5.3) & 21.8 \\ \hline


\textbf{Gender} \\ \hline
Female & 8,090 (44.7) & 13.5 \\ \hline
Male & 10,004 (55.3) & 13 \\ \hline

\textbf{Insurance} \\ \hline
Medicare & 10,337 (57.1) & 15.3 \\ \hline
Medicaid & 1,489 (8.2) & 10.3 \\ \hline
Private & 5,601 (31.0) & 10.2 \\ \hline
Other & 667 (3.7)  & 11.6 \\ \hline

\textbf{Marital Status} \\ \hline
Married & 8,564 (45.3) & 46.4 \\ \hline
Single & 4,422 (24.4) & 19.3 \\ \hline      
Widowed & 2,654 (14.7) & 17.4 \\ \hline     
Divorced & 1,119 (6.2) & 5.4 \\ \hline
Separated &  194 (1.0) & 1.1 \\ \hline       
Unknown (default) & 302 (1.7) &  1.1 \\ \hline
Life partner & 5 (0.03)  & 0 \\ \hline

\end{tabular}
\label{mimic_demo_dist}
\end{table}
\end{singlespace}

With this dataset, we predicted in-hospital mortality (IHM) as a binary label. We determined IHM by comparing the patient's date of death with hospital admission and discharge times. The resulting mortality rate is 13.23\% (2,797 of 21,139 ICU stays). We processed data by reproducing the method in \cite{harutyunyan2019multitask}. A full description can be found in the original publication. To process the data for fairness evaluation, we binary-encoded each protected label and assigned them as privileged and unprivileged classes to analyze them using our chosen fairness metrics. We encoded all participants with age $<=60$ as the unprivileged class (0), and those $>60$ as the privileged class (1). The reasoning behind this encoding is that the Centers for Disease Control and Prevention (CDC) defines an “older adult” as someone who is at least 60 years old (\cite{centers2012identifying}). For the Gender label, male was encoded as the privileged class (1) due to its higher number of samples and because female patients appear to have a slightly higher IHM risk, making the female class the unprivileged class (0). We classified the participants according to whether they have private or government-assisted insurance. Those with private or self-care insurance represented the privileged class (1), and those with Medicare, Medicaid, and Government represented the unprivileged class (0). The reasoning behind this is the connection between government-aided insurance and socioeconomic conditions and age as mentioned earlier in this section. For marital status, we categorized married patients, those with life partners as privileged, and all others (no partner) as unprivileged.  We did this in accordance with research
showing that unmarried patients, including those who are divorced, separated, widowed, or never married, show elevated rates of certain health complications and death compared to those with partners (\cite{dhindsa2020marital, johnson2000marital}).

\subsection{SNAPSHOT Data Description}

The SNAPSHOT dataset includes multi-modal physiological, behavioral, and survey data collected from 350 college students in one university, totaling over 7500 days of data (\cite{taylor2017personalized}). The study was conducted between 2013 and 2017, within which 30 days of consecutive data were recorded for 228 participants and 90 days for 15 participants. The data recorded includes wrist-worn wearable sensor data (acceleration, skin conductance, and skin temperature), mobile phone data (call, SMS, screen on/off logs, GPS locations), weather data (obtained using DarkSky API \cite{dark2015dark}), self-reported daily morning and evening well-being (non-numeric scales or mood, health, and stress later scored 0-100), genders, and Big Five Personality scores (\cite{john1999big}). 

From the SNAPSHOT data, we computed a total of 378 features, including 173 physiological features that include Electrodermal activity (EDA) features, skin conductance Level (SCL), skin temperature features, accelerometer features, 165 mobile phone features, 43 weather features from the raw data, and statistical features calculated from each of these classes of features. Tables \ref{snap_sleep_feature_description}, \ref{snap_activity_feature_description}, \ref{snap_sensor_feature_description}, \ref{snap_sensor_feature_description2}, \ref{snap_weather_feature_description}, and \ref{snap_weather_feature_description2} summarize these features. For more details on the features extracted, refer to \cite{taylor2017personalized}. 

\begin{singlespace}
\begin{table}[htbp]
\centering
\caption{SNAPSHOT: Description of Sleep and Nap Related Survey Data}
\begin{tabularx}{\textwidth}{|l|X|}

\hline
\textbf{Features} & \textbf{Description} \\ \hline

State Score & State anxiety score\\ \hline
no\_sleep\_24 & Number of times participant slept in 24 hours. \\ \hline
sleep\_latency & Time taken to fall asleep\\ \hline
pre\_sleep\_activity & What activity the participant performed before going to sleep. \\ \hline
awakening & If the participant woke up through the night (yes or no) \\ \hline
awakening\_occations & Number of awakenings\\ \hline
wake\_reason & How the participant woke up (spontaneously, alarm, or disturbance). \\ \hline
count\_awakening & How many times participant woke up through the night. \\ \hline
awakening\_duration &  How long was the participant awake for if they woke up through the night. \\ \hline
nap & If the participant took a nap (yes or no) \\ \hline
nap\_occations & Number of naps \\ \hline
count\_nap & How many times the participant took a nap. \\ \hline
nap\_duration & How long the nap was. \\ \hline
first\_event\_none & If an event is scheduled that day (yes or no)\\ \hline
time\_in\_bed & How long the participant spent in bed for the day. \\ \hline
sleep\_try\_time\_mins\_since\_midnight & What time the participant tries to go to sleep \\ \hline
wake\_time\_mins\_since\_midnight & Wake time in minutes since midnight\\ \hline
first\_event\_mins\_since\_midnight & first event time in minutes since midnight\\ \hline
presleep\_media\_interaction & If the participant has presleep media interaction (yes or no) \\ \hline
presleep\_personal\_interaction & If the participant has presleep personal interaction (yes or no) \\ \hline
positive\_interaction & If the participant had any positive interactions with someone for the day (yes or no). \\ \hline
negative\_interaction & If the participant had any negative interactions with someone for the day (yes or no). \\ \hline

\end{tabularx}
\label{snap_sleep_feature_description}
\end{table}

\begin{table}[htbp]
\centering
\caption{SNAPSHOT: Description of Activity Related Survey Data}
\begin{tabularx}{\textwidth}{|l|X|}

\hline
\textbf{Features} & \textbf{Description} \\ \hline

academic & If the participant attended any academic activities (yes or no). \\ \hline
count\_academic & How many academic activities the participant attended. \\ \hline
academic\_duration  & How long the participant spent on academic activities \\ \hline
study\_duration & How many hours participant studied for, outside of academic activities. \\ \hline
exercise & If the participants engaged in any exercise-based activities (yes or no) \\ \hline
exercise\_occations  & How many times the participant engaged in exercise-based activities. \\ \hline
exercise\_duration  & For how long the participant exercised. \\ \hline
extracurricular & If the participant attended any other extracurricular activities (yes or no). \\ \hline
count\_extracurricular & How many extracurricular activities the participant attended in the day. \\ \hline
extracurricular\_duration  & How long the participant attended extracurricular activities for. \\ \hline
overslept & If the participant overslept and missed any scheduled events.\\ \hline
caffeine\_count & Total servings of caffeine participant had for the day. \\ \hline
drugs & If the participant had any other drugs or medication besides  (yes or no). \\ \hline
drugs\_alcohol & If the participant had any alcohol (yes or no). \\ \hline
drugs\_alert & If the participant had any drugs to keep them alert (yes or no). \\ \hline
drugs\_sleepy & If the participant had any drugs that made them sleepy (yes or no). \\ \hline
drugs\_tired & If the participant had any drugs that made them tired (yes or no).\\ \hline

\end{tabularx}
\label{snap_activity_feature_description}
\end{table}

\begin{table}[htbp]
\centering
\caption{SNAPSHOT: Description of Electrodermal Activity and Skin Conductance Features Extracted from Wearable Sensor Data}
\begin{tabularx}{\textwidth}{|X|X|}

\hline
\textbf{Features} & \textbf{Description} \\ \hline

\textbf{Electrodermal activity (EDA) Peak Features} \\ \hline

Sum AUC & Sum of the AUC of all peaks for this period where
the amplitude of the peak is calculated as the difference
from base tonic signal \\ \hline

Sum AUC Full & Sum of AUC of peaks where amplitude is calculated as difference from 0 \\ \hline

Median RiseTime & Median rise time for peaks (seconds) \\ \hline

Median Amplitude & Median amplitude of peaks ($\mu$S) \\ \hline

Count Peaks & Number of detected peaks \\ \hline

SD Peaks 30 min & Compute number of peaks per 30 minute epoch, take standard deviation of this signal \\ \hline

Med Peaks 30 min & Compute number of peaks per 30 minute epoch, take median of this signal \\ \hline

Percent Med Peak & Percentage of signal containing 1 minute epochs with
greater than 5 peaks \\ \hline

Percent High Peak & Same as Percent Med Peak \\ \hline

\textbf{Skin  Conductance Level (SCL) Features} \\ \hline

Percent Off & Percentage of period where sensor was off \\ \hline

Max Unnorm & Maximum level of un-normalized EDA signal \\ \hline

Med Unnorm & Median of un-normalized EDA signal \\ \hline

Mean Unnorm & Mean of un-normalized EDA signal \\ \hline

Median Norm & Median of z-score normalized EDA signal \\ \hline

SD Norm & Standard Deviation of z-score normalized EDA signal \\ \hline

Mean Deriv & Mean derivative of z-score normalized EDA signal ($\mu$S/second) \\ \hline

\end{tabularx}
\label{snap_sensor_feature_description}
\end{table}

\begin{table}[htbp]
\centering
\caption{SNAPSHOT: Description of Accelerometer and Skin Temperature Features Extracted from Wearable Sensor Data}
\begin{tabularx}{\textwidth}{|l|X|}

\hline
\textbf{Features} & \textbf{Description} \\ \hline

\textbf{Accelerometer Features} \\ \hline

Step Count & Number of steps detected \\ \hline

Mean Movement Step Time & Average number of samples (at 8Hz) between two steps (aggregated first to 1 minute, then we take the mean of only the
parts of this signal occurring during movement) \\ \hline

Stillness Percent & Percentage of time the person spent nearly motionless \\ \hline

Sum Stillness weighted AUC & Sum the weights of the peak AUC signal by how still the user was every 5 minutes \\ \hline

Sum Steps Weighted AUC & Sum the weights of the peak AUC signal by the step count over every 5 minutes \\ \hline

Sum Stillness Weighted Peaks & Multiply the number of peaks every 5 minutes by the
amount of stillness during that period \\ \hline

Max Stillness Weighted Peaks & Max value for the number of peaks multiplied by the  stillness for any five minute period \\ \hline

Sum Steps Weighted Peaks & Divide number of peaks every five minutes by step count and sum \\ \hline

Med Steps Weighted Peaks & Average value of number of peaks divided by step count for every 5 mins \\ \hline

\textbf{Skin Temperature (ST) Features} \\ \hline

Max Raw Temp & Maximum of the raw temperature signal (\degree C) \\ \hline

Min Raw Temp & Minimum of the raw temperature signal (\degree C) \\ \hline

SD Raw Temp & Standard deviation of the raw temperature signal \\ \hline

Med Raw Temp & Standard deviation of the raw temperature signal \\ \hline

Sum Temp Weighted AUC & Sum of peak AUC divided by the average temp for every 5 minutes \\ \hline

Sum Temp Weighted Peaks & Number of peaks divided by the average temp for every 5 minutes \\ \hline

Max Temp Weighted Peaks & Maximum number of peaks in any 5 minute period
divided by the average temperature \\ \hline

SD Stillness Temp & Standard deviation of the temperature recorded during periods when
the person was still \\ \hline

Med Stillness Temp & Median of the temperature recorded during periods when
the person was still \\ \hline

\end{tabularx}
\label{snap_sensor_feature_description2}
\end{table}

\begin{table}[htbp]
\centering
\caption{SNAPSHOT: Description of Features Extracted from Weather API 1}
\begin{tabularx}{\textwidth}{|l|X|}

\hline
\textbf{Features} & \textbf{Description} \\ \hline

Sunrise & Time of sunrise (UTC) \\ \hline
Moon\_phase & Moon phase value on a scale of 0−1(new moon-full moon) \\ \hline
Apparent\_temp\_max & Maximum apparent temperature of the day (\textdegree F) \\ \hline
Apparent\_temp\_min & Minimum apparent temperature of the day (\textdegree F) \\ \hline
Temperature\_max & Maximum temperature of the day (\textdegree F) \\ \hline
Temperature\_min & Minimum temperature of the day (\textdegree F) \\ \hline
Avg\_cloud\_cover & Percentage of sky covered by cloud on a scale of 0-1 \\ \hline
Avg\_dew\_point & Average dew point temperature \\ \hline
Avg\_humidity & Daily average value of humidity on a scale of 0-1 \\ \hline
Avg\_pressure & Average atmospheric pressure on the sea level (hPa) \\ \hline
Morning\_pressure\_change & Trinary value of pressure difference between midnight and noon (rising, falling, steady) \\ \hline
Evening\_pressure\_change & Trinary value of pressure difference between noon and midnight (rising, falling, steady) \\ \hline
Avg\_visibility & Average visibility (meters) \\ \hline
weather\_precip\_probability & Precipitation probability \\ \hline
Temperature\_rolling\_mean & Rolling average of temperature \\ \hline
Temperature\_rolling\_std & Rolling standard deviation of temperature \\ \hline
apparentTemperature\_rolling\_mean & Rolling average of apparent temperature \\ \hline
apparentTemperature\_rolling\_std & Rolling standard deviation of apparent temperature \\ \hline
apparentTemperature\_today\_vs\_avg\_past & Difference between today’s apparent temperature and its rolling average \\ \hline
pressure\_rolling\_mean & Rolling average of pressure \\ \hline
pressure\_rolling\_std & Rolling standard deviation of pressure \\ \hline
apparentTemperature\_today\_vs\_avg\_past & Difference in today’s apparent temperature and its rolling average \\ \hline
pressure\_rolling\_mean & Rolling average of pressure \\ \hline
pressure\_rolling\_std & Rolling standard deviation of pressure \\ \hline
pressure\_today\_vs\_avg\_past & Difference between today’s pressure and its rolling average \\ \hline
cloudCover\_rolling\_mean & Rolling average of cloud cover \\ \hline
cloudCover\_rolling\_std & Rolling standard deviation of cloud cover \\ \hline
cloudCover\_today\_vs\_avg\_past & Difference between today’s cloud cover and its rolling average \\ \hline
humidity\_rolling\_mean & Rolling average humidity \\ \hline
humidity\_rolling\_std & Rolling standard deviation humidity \\ \hline
humidity\_today\_vs\_avg\_past & Difference between today’s humidity and its rolling average \\ \hline

\end{tabularx}
\label{snap_weather_feature_description}
\end{table}

\begin{table}[htbp]
\centering
\caption{SNAPSHOT: Description of Features Extracted from Weather API 2}
\begin{tabularx}{\textwidth}{|l|X|}

\hline
\textbf{Features} & \textbf{Description} \\ \hline

windSpeed\_rolling\_mean & Rolling average of wind speed \\ \hline
windSpeed\_rolling\_std & Rolling standard deviation of wind speed \\ \hline
windSpeed\_today\_vs\_avg\_past & Difference between today’s wind speed and its rolling average \\ \hline
precipProbability\_rolling\_mean & Rolling average of precipitation probability \\ \hline
precipProbability\_rolling\_std & Rolling standard deviation of precipitation probability \\ \hline
precipProbability\_today\_vs\_avg\_past & Difference between current precipitation probability and its rolling average \\ \hline
sunlight & Duration of sunlight (sec) \\ \hline
quality\_of\_day & Quality of the day defined in terms of 8 categories in the range {−4, 4}: clear=4, partly-cloudy=3, cloudy=2, wind=1, fog=-1, rain=-2, sleet=-3, snow=-4 \\ \hline
avg\_quality\_of\_day & Average value for quality\_of\_day \\ \hline
precipType & Type of precipitation as integer: None=0, Rain=1, Hail=2, Sleet=3, Snow=4, Other=5 \\ \hline
max\_precip\_intensity & Maximum Precipitation volume (mm) \\ \hline
median\_wind\_speed & Median wind speed of the day (meter/sec) \\ \hline
median\_wind\_bearing & Median wind bearing of the day (degrees) \\ \hline

\end{tabularx}
\label{snap_weather_feature_description2}
\end{table}

\end{singlespace}

From the SNAPSHOT data, we predicted 4 self-reported mood labels, namely morning and evening happiness and calmness levels, using physiological features, mobile phone usage, and weather. We utilized the previous seven (7) days of data to predict the current day's label. These labels appear as a score of 1 to 100 in the data set, and we binarized them with all scores less than or equal to 50 to mean sad or stressed and above 50 means happy or calm. 

\begin{figure*}[htbp]
\includegraphics[width=\textwidth]{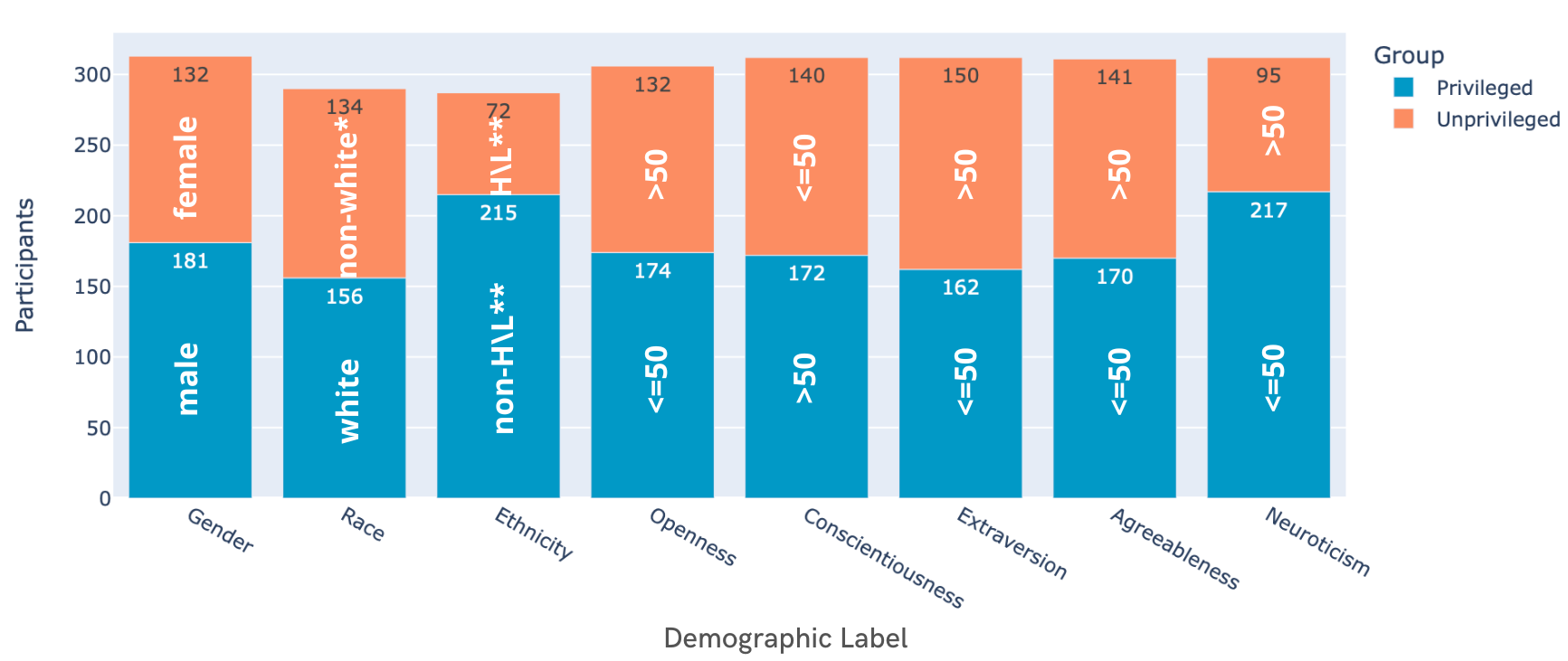}
\caption{SNAPSHOT Demographic Distribution}
\label{fig-snap-demo}
\end{figure*}

The SNAPSHOT dataset contains numerous demographic information on the participants, and we used 3 of these: gender, race, and ethnicity as protected labels. For these labels, we chose the class with more participants as the privileged class, following the concept of negative legacy or data bias (\cite{cunningham2021underestimation}). We also used personality types, including openness, conscientiousness, extraversion, agreeableness, and neuroticism, as protected labels in our experiments. Figure \ref{fig-snap-demo} shows a distribution of all the protected labels across this dataset, where non-white refers to all other races other than white, which include Asian (77 participants), Black or African American (38 participants), American Indian or Alaskan Native (6), and other (13 participants). H\textbackslash L refers to Hispanic and Latino in the figure.

Some participants in the SNAPSHOT dataset do not have race and ethnicity information, and we dropped samples for such participants when running experiments for race and ethnicity labels. In the end, we dropped 314 samples from the training data, reducing it from 4030 to 3716 samples, and the test samples were reduced from 1866 to 1715. We split the data into 70\% training and 30\% testing sets for all experiments.

\section*{Appendix B.Experiments}

In this appendix, we provide more information on the experimental process from Section 5. We provide details on baseline model implementations, model structures and parameters, and elaborate on the Reweighting method.

\subsection*{Baseline Model Implementations}

For the ADULT dataset, we built a Convolutional Neural Network (CNN) with 2 dense layers and a dropout later with a dropout rate of 0.25 to predict binary income labels; $<=50$k as 0 and $>50$k as 1. We used a 3-fold cross-validation method and binary cross entropy as the loss function with an Adam optimizer. The structure of this model is shown below.

For the MIMIC-III dataset, we used a Long-Short Term Memory (LSTM)-based model proposed in \cite{harutyunyan2019multitask}. LSTM is a Recurrent Neural Network (RNN) which is a type of Artificial Neural Network (ANN) designed to capture long-term dependencies in sequential data (\cite{harutyunyan2019multitask}). To implement our model, we resampled the time series data into regularly spaced intervals. In the case of multiple measurements of the same variable in the same interval, we utilized the value of the last measurement. We used a 48-hour window for prediction, enabling the detection of patterns that may indicate changes in patient acuity, as proposed by \cite{harutyunyan2019multitask}. We selected the channel-wise LSTM model without deep supervision for our study based on its superior reported area under the receiver operating characteristic (AUROC) performance among the five developed non-MTL models by \cite{harutyunyan2019multitask}. The selected network consists of a bi-directional layer, an LSTM layer with 16 neurons, and a dropout layer with a dropout rate of 0.3 (structure shown in Appendix B). We used a batch size of 8, binary cross entropy as the loss function and Adam optimizer. We executed this for 100 epochs, reproducing the experiments from \cite{harutyunyan2019multitask}.

The baseline model we used for predicting the labels for the SNAPSHOT dataset is a gated recurrent unit (GRU) model with 2 GRU layers and 2 dropout layers with a dropout rate of 0.5, 30 neurons and focal loss with a gamma value of 2, and an alpha value of 4. This model predicts 4 labels, namely: "morning happiness", "morning calmness", "evening happiness", and "evening calmness". We trained this model for 300 epochs with a batch size of 1000 and a focal loss function since there is a significant class imbalance in the target label for this dataset. Focal loss is an extension of the cross-entropy loss function that would down-weight easy examples and focus training on hard negatives (\cite{lin2017focal}). It reduces the loss contribution from easy examples and increases the importance of correcting misclassified examples. The structure of this model is also shown in Appendix B. With 8 protected labels, which are gender, race, ethnicity, openness, conscientiousness, extraversion, agreeableness, and neuroticism, and 4 target labels, we conducted 32 sets of experiments to analyze the bias in this dataset.

\subsection*{Model Structures}

\begin{figure*}[h]
\begin{center}
\includegraphics[scale=0.28]{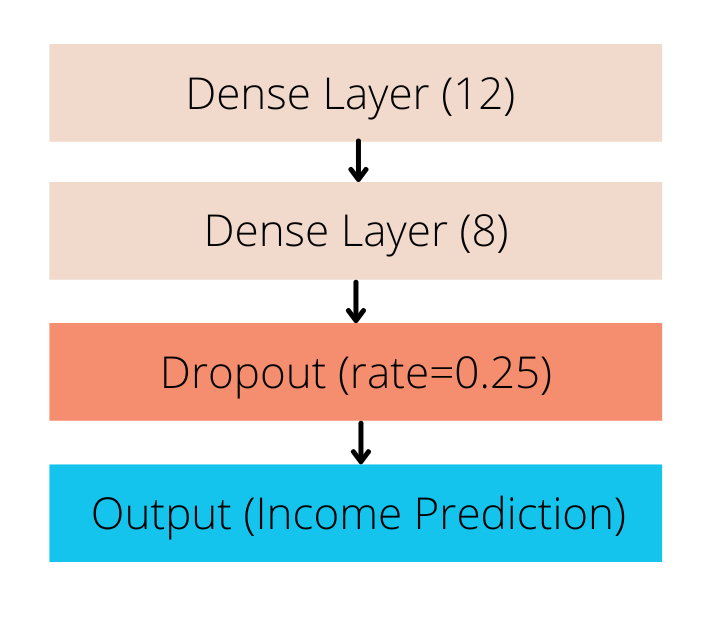}
\caption{ADULT: Income Prediction Model Structure}
\label{fig-adult-model}
\end{center}
\end{figure*}

\begin{figure*}[h]
\begin{center}
\includegraphics[scale=0.28]{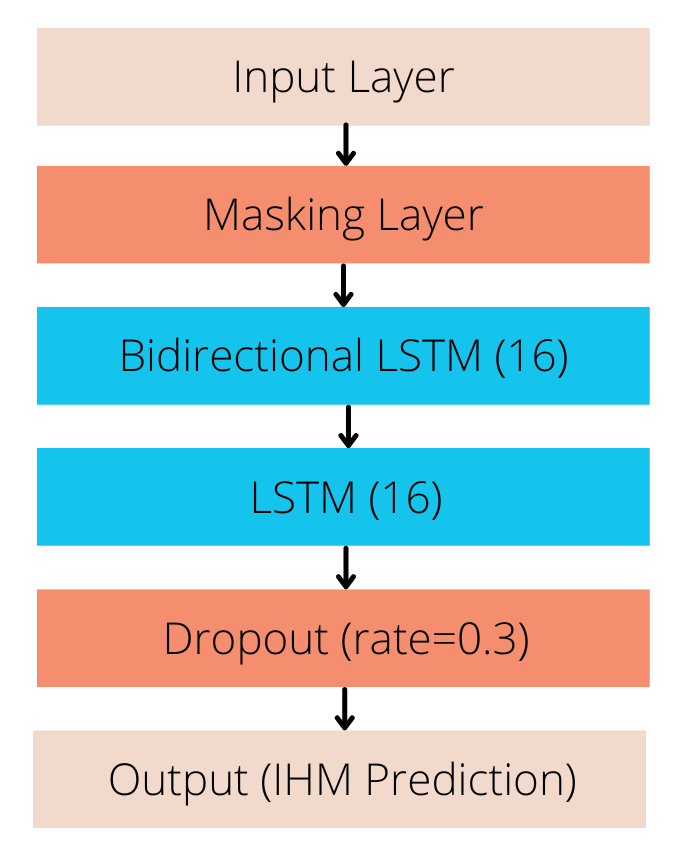}
\caption{MIMIC-III: IHM prediction Model Structure}
\label{fig-mimic-model}
\end{center}
\end{figure*}

\begin{figure*}[htbp]
\begin{center}
\includegraphics[scale=0.28]{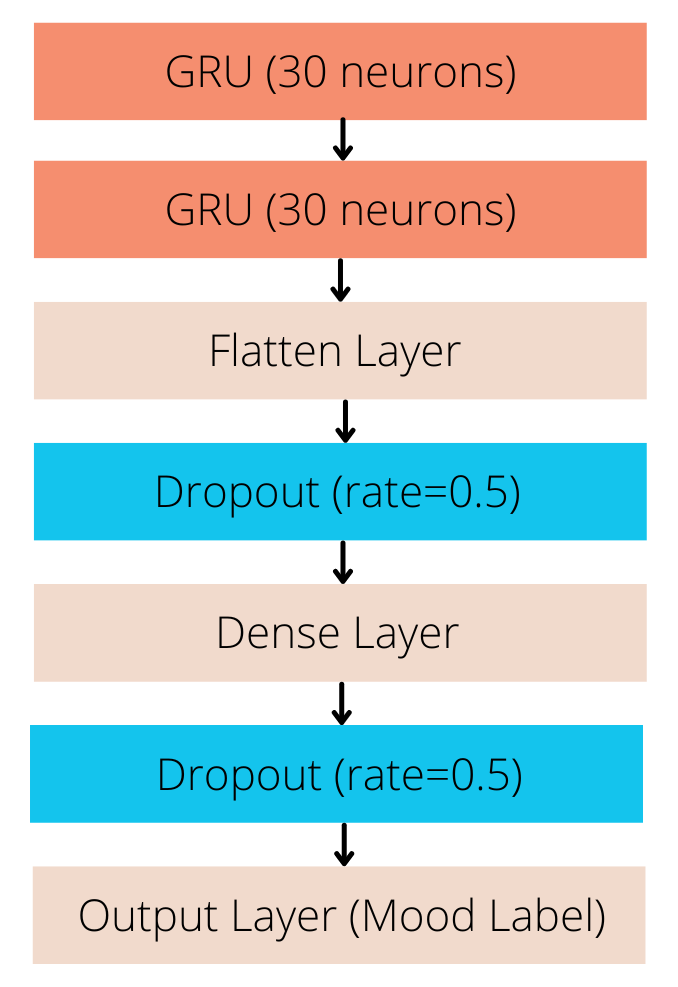}
\caption{SNAPSHOT: Mood Labels Prediction Model Structure}
\label{fig-snap-model}
\end{center}
\end{figure*}

\subsection*{Proposed Method Implementation Details}

We applied our bias-reducing MTL method by developing MTL models to predict target and protected labels by which the baseline models are biased. We first trained the MTL model for each of the combinations for all datasets. We assigned different loss weight ratios to the label pairs, prioritizing the prediction label over the protected label to ensure that the model prioritizes the prediction label and make certain that its uncertainty score is kept lower than the protected label prediction. For the ADULT dataset, the ratio was 4.5 to 0.25; for MIMIC-III, 5 to 0.5; and for SNAPSHOT, 6 to 0.25. We arrived at these combinations by experimenting with a few different combinations. For example, with the ADULT dataset, the uncertainty of the age label remained around 0.60 for varying ratios up to 4 to 1 and improved to 0.62 at ratios 4 to 0.25. For the SNAPSHOT dataset, we used Focal loss as the loss function for both predictions and assigned gamma values 2 and 0 for the chosen target and protected labels, respectively. We did this because setting gamma\>0 reduces the relative loss for well-classified examples. We trained the MTL networks for 100 epochs for all the datasets and saved the weights at each epoch, making it a total of 100 weight combinations for each experiment run.

\subsection*{Data Manipulation on SNAPSHOT Dataset}

To create a more significant imbalance for the gender label in the SNAPSHOT dataset, we manipulated the data in the following way:

Originally, the data contained samples from 181 male and 132 female participants. There was a total of 4030 train and 1866 test samples, and we removed the samples from 75 participants that identified as female from the training data, making it a total of 2660 samples for training. We removed participants with 14 or more samples in the train data (experiment S-1). We also experimented with removing samples from participants with 5 or more samples in the train data (experiment S-2) and those with 30 or more samples (experiment S-3), leaving 2349 and 3685 samples, respectively. The motivation is to induce an artificial data bias, otherwise known as a negative legacy, into our model by creating a higher imbalance in the representation of male and female participant samples in our training data. With this manipulation, we expect our model to not generalize well on the test data that still contains a comparable number of female samples as hypothesized by \cite{gorrostieta2019gender}, thereby inducing some bias in its predictions.

\subsection*{Reweighting} \label{sup_reweighting}

We compare our proposed method to a standard bias mitigation method called Reweighing introduced by \cite{kamiran2012data}. Reweighing is a pre-processing bias mitigation technique. It involves applying appropriate weights to different tuples in the training dataset to make it discrimination free with respect to the protected label  (\cite{park2021comparison}). It generates weights for the training examples in each (group, label) combination differently to ensure fairness before classification (\cite{bellamy2018ai}). See the details of reweighting in Appendix B. 

We apply the AI Fairness 360 implementation of this method to our processed data for all datasets and pass the weight vectors to our baseline model for all 3 datasets and calculated their fairness metric scores to compare to our proposed method.

The Reweighting bias mitigation technique works by applying different weights to each group-label combination according to the conditional probability of label by the protected label. It assigns different weights to various subgroups, determined by the relationship between the protected label and the outcome label. For example, if certain groups are underrepresented in positive outcomes, Reweighting increases their influence in the training process by assigning higher weights to these groups. This process aims to equalize the predictive performance across groups, mitigating any potential bias that the model may learn from imbalanced or skewed data.

Assume T and "Label" represent the protected and target labels simultaneously. Samples with T=t and Label = + will get higher weights than samples with T=t and Label = -, and samples with T$\neq$t and Label = + will get lower weights than samples with T$\neq$t and Label = -, where T represents the protected label, and Label represents the target label. According to these weights, the objects will be sampled (with replacement), leading to a dataset without dependency (\cite{calders2009building}). These weights are calculated as follows using some basic notions of probability theory.

Assuming the dataset D is unbiased, in the sense that T and Label are independent, the expected probability $P_{exp}(t\wedge+)$ would be:

$$P_{exp}(t\wedge+):=t\times+$$

where t is the fraction of objects having T=t and ‘+’ the fraction of tuples having Label == +. In reality, however, the actual probability

$$P_{act}(t\wedge+):=t\wedge+$$

might be different. If the expected probability is higher than the actual probability value, it shows the bias towards label ‘-’ for T=t. We assign weights to t with respect to label ‘+’. The weight will be:

\begin{equation*}
W(T=t\vert x(Label)=+): ={P_{exp}(t\wedge+) \over P_{act}(t\wedge+)}
\label{eq:reweighing_1}
\end{equation*}

This weight of t for label '+' will over-sample objects with T=t for the label '+'. The weight of t for label '-' becomes:

\begin{equation*}
W(T=t\vert x(Label)=-): ={P_{exp}(t\wedge-) \over P_{act}(t\wedge-)}
\label{eq:reweighing_2}
\end{equation*}


When applied to for example the ADULT dataset taking sex as the protected label, to remove the dependency between sex (T) and the target label (Label), we calculate a weight for each data object according to its T and Label value. In this example where both T=sex and Label are both binary attributes, only 4 combinations between the values of T and Label are possible, i.e., T=female(f) or T=male(m) can have Label values '+' or '-'. We can then calculate the weight of a data sample with T=f and Label '+'. Let's assume that approximately 50\% samples have T=m and 60\% have Label value '-'. so the expected probability of the sample can be computed as:

$$P_{exp}(Sex=m |x(label)=-)=0.50 \times 0.60 $$

but its actual probability is 20\%. So the weight W will be:

\begin{equation*}
W(Sex=m | x(Label)=-) = {0.50 \times 0.60 \over 0.2} = 1.5
\label{eq:reweighing_3}
\end{equation*}

The weights for the other combinations are also calculated in a similar manner. 

We assign to every tuple a weight according to its T and Label values. The balanced dataset is then created by sampling the original training data, replaced with the assigned weights. On this balanced dataset the dependency-free classifier is learned. The Reweighing technique can be seen as an instance of cost-sensitive learning (\cite{elkan2001foundations}) in which, e.g., an object of label ‘+’ with T=t gets a higher weight, making an error for this object more "expensive". The pseudocode of the algorithm describing this Reweighing technique can be found in \cite{calders2009building} and \cite{kamiran2009classifying}.

\subsubsection*{Other Baseline Methods}

We compare our proposed method's performance on the ADULT and MIMIC-III datasets against the Fairness with Related Features (FairRF) method, introduced by \cite{zhao2022towards}. The FairRF develops fair classifiers by leveraging features related to but not explicitly revealing sensitive attributes. The method focuses on minimizing the correlation between these related features and model predictions to ensure fairness. FairRF adjusts the importance weights of each feature and has shown effectiveness in enhancing classifier fairness while maintaining high accuracy across various real-world datasets.

Additionally, we review a fair algorithm called Adversarial Reweighted Learning (ARL) introduced by \cite{lahoti2020fairness} against our results for MIMIC-III. This approach focuses on identifying and mitigating biases that arise from non-protected features that are often correlated with sensitive attributes. The core idea of ARL is to use adversarial training techniques to re-weight the importance of features during the learning process. The adversarial component involves a dual-training setup where one model aims to predict the target outcome based on re-weighted features, while a second adversarial model attempts to predict the protected attribute from the same features. The goal is to adjust the weights of features in a way that the primary model maintains high accuracy on the target task, while the adversarial model fails to predict the protected attribute, thus ensuring that the predictions are fair and unbiased. This method was tested across various datasets and has shown to improve fairness metrics significantly without substantial loss in performance.

\section*{Appendix C. Saliency Maps for MIMIC-III and SNAPSHOT Datasets}

\begin{figure*}[htbp]
\centerline{\includegraphics[scale=0.4]{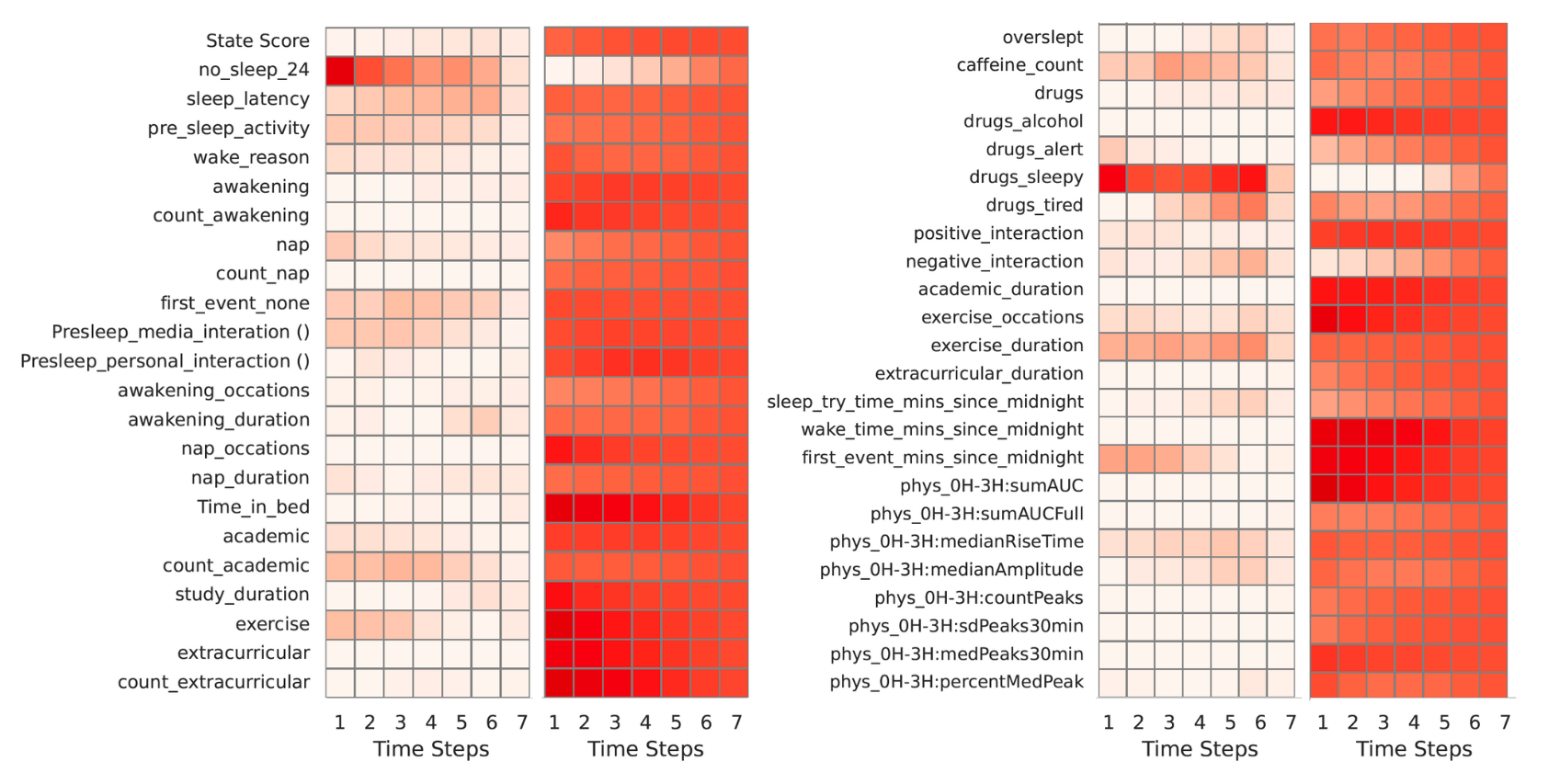}}
\caption{SNAPSHOT: Comparison of Saliency Maps From Baseline Model and Model After Proposed Method Implementation for evening-sad-happy Label Prediction for 47 features-1.}
\label{fig-snap_sal_1}
\end{figure*}

\begin{figure*}[htbp]
\centerline{\includegraphics[scale=0.4]{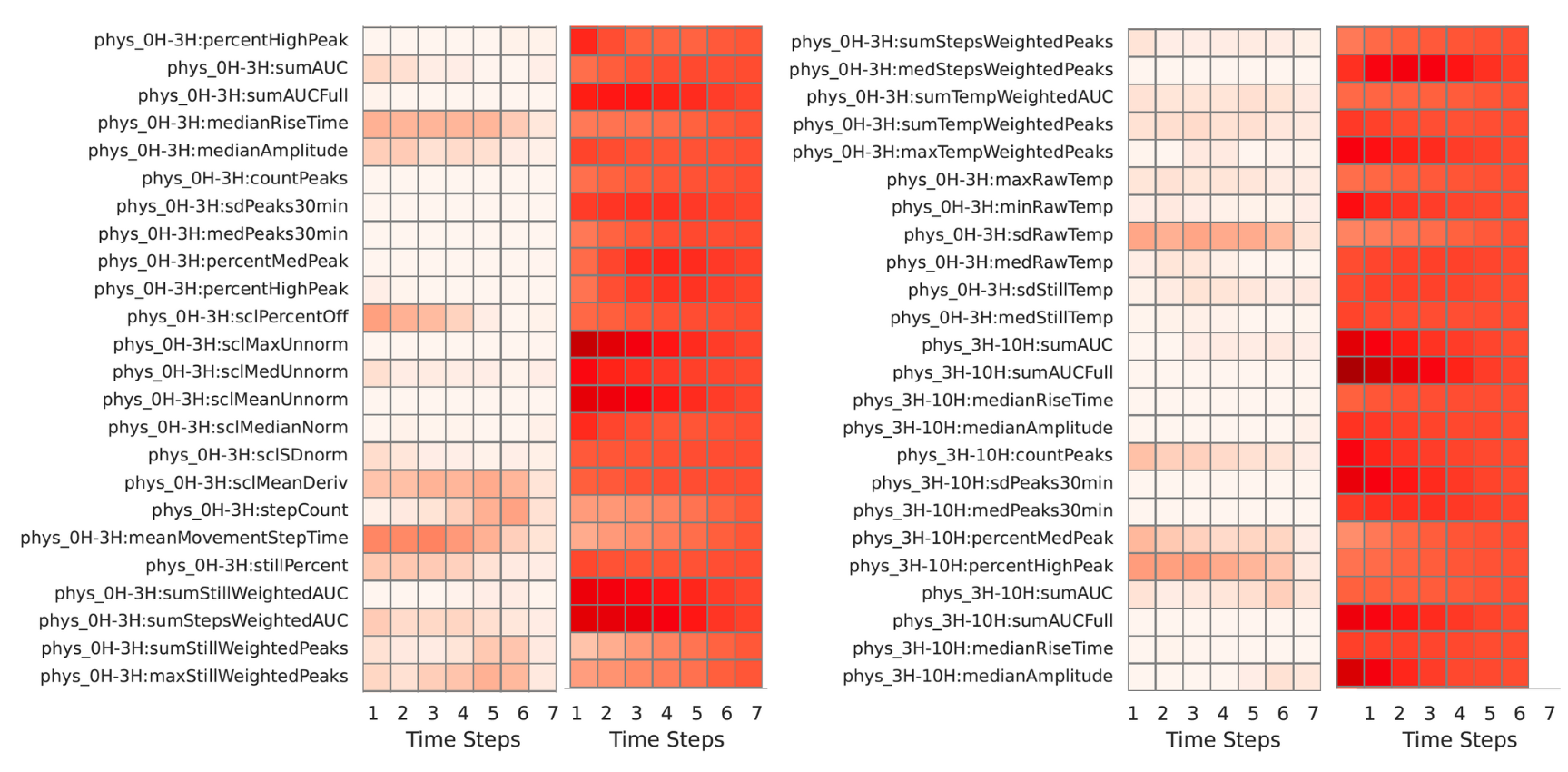}}
\caption{SNAPSHOT: Comparison of Saliency Maps From Baseline Model and Model After Proposed Method Implementation for evening-sad-happy Label Prediction for 48 features-2.}
\label{fig-snap_sal_2}
\end{figure*}

\begin{figure*}[htbp]
\centerline{\includegraphics[scale=0.4]{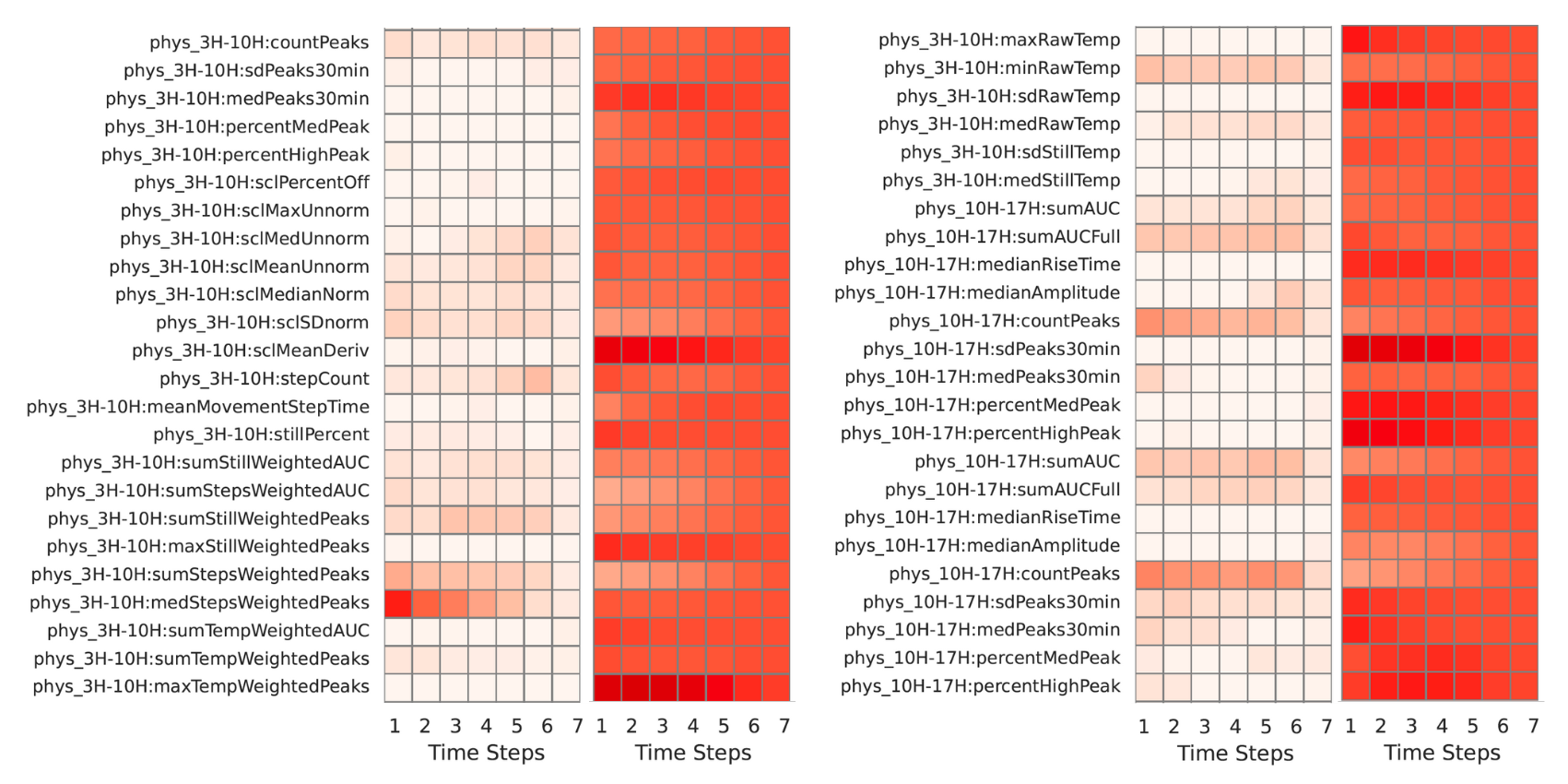}}
\caption{SNAPSHOT: Comparison of Saliency Maps From Baseline Model and Model After Proposed Method Implementation for evening-sad-happy Label Prediction for 48 features-3.}
\label{fig-snap_sal_3}
\end{figure*}

\begin{figure*}[htbp]
\centerline{\includegraphics[scale=0.4]{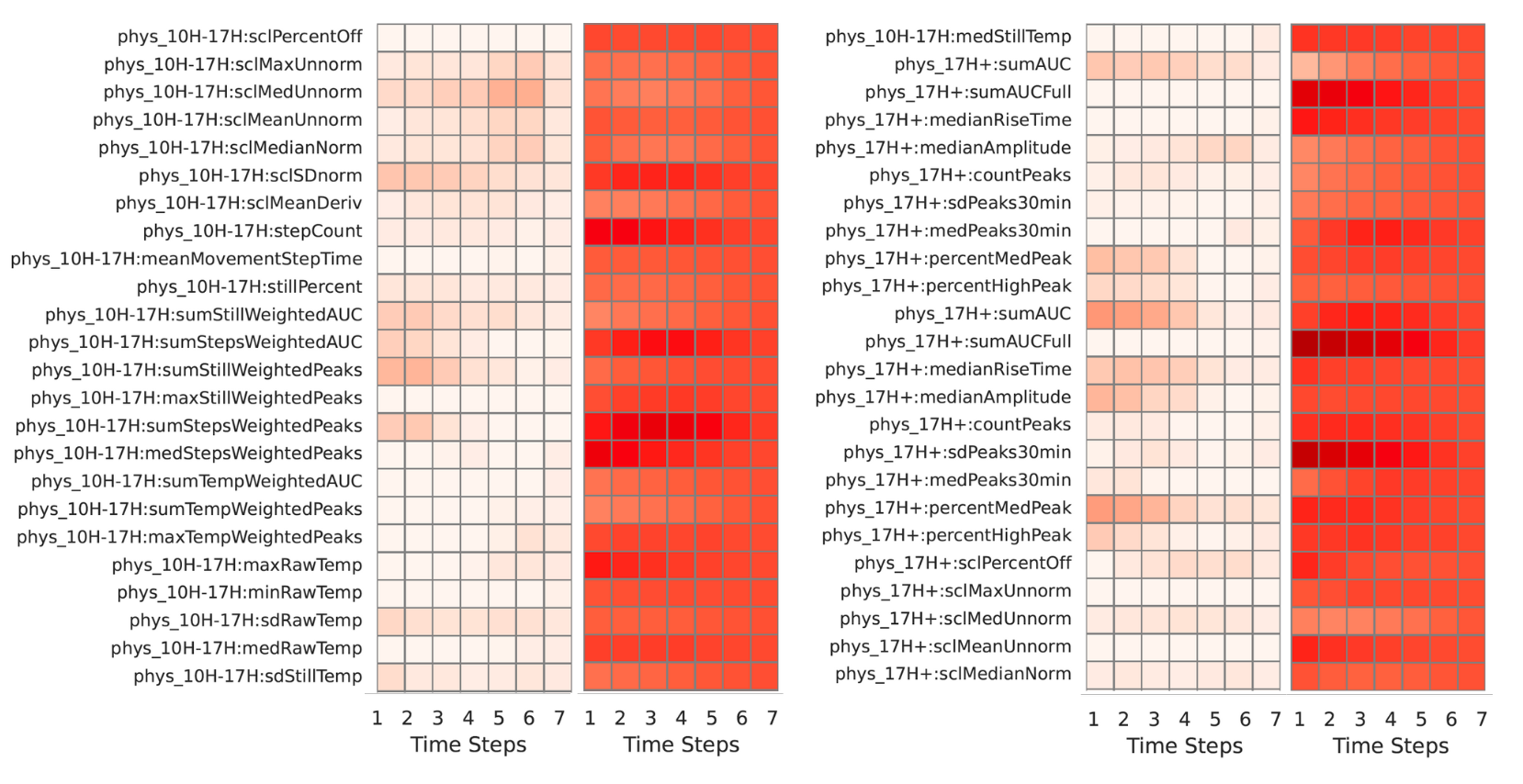}}
\caption{SNAPSHOT: Comparison of Saliency Maps From Baseline Model and Model After Proposed Method Implementation for evening-sad-happy Label Prediction for 48 features-4.}
\label{fig-snap_sal_4}
\end{figure*}

\begin{figure*}[htbp]
\centerline{\includegraphics[scale=0.4]{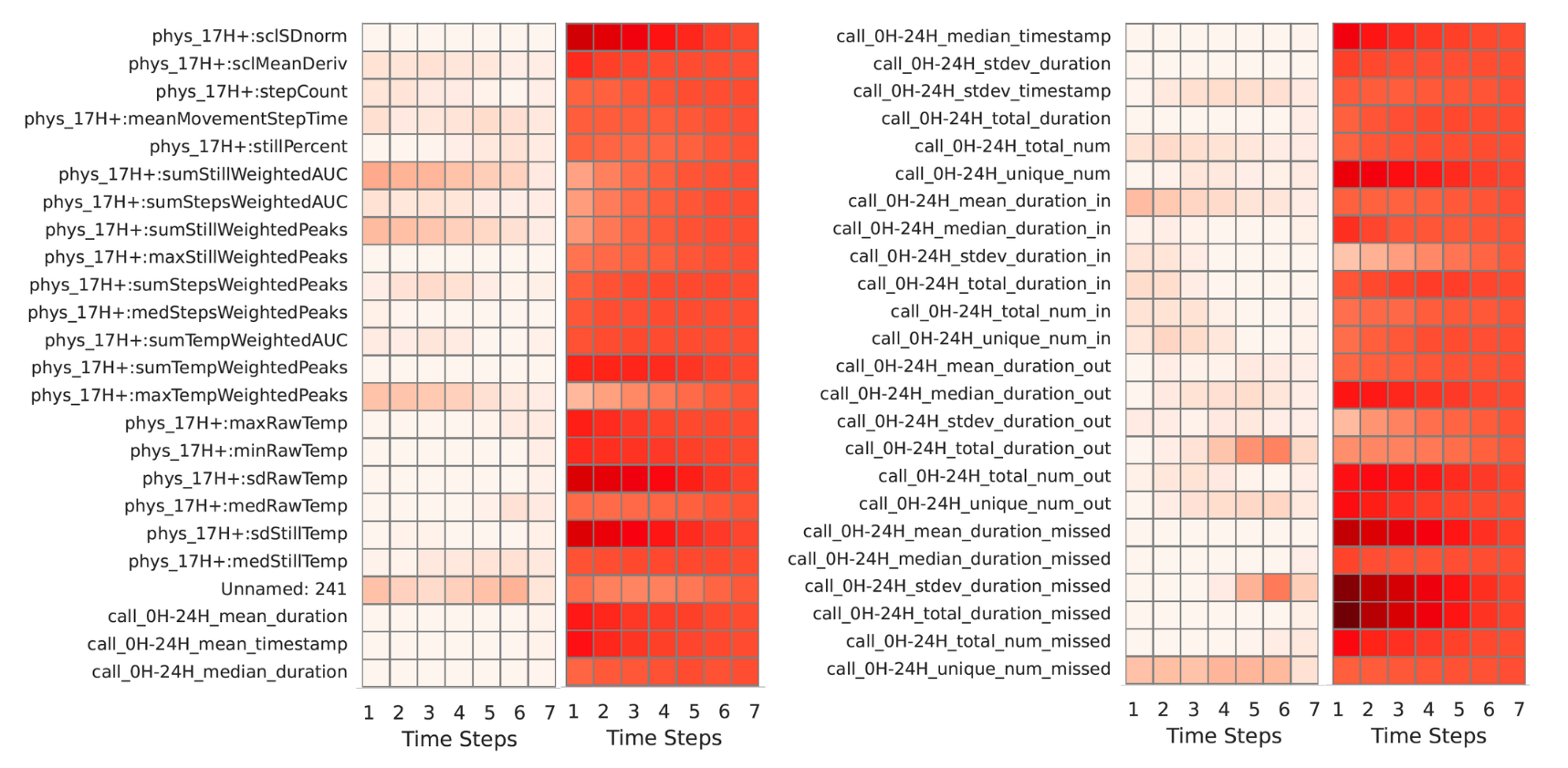}}
\caption{SNAPSHOT: Comparison of Saliency Maps From Baseline Model and Model After Proposed Method Implementation for evening-sad-happy Label Prediction for 48 features-5.}
\label{fig-snap_sal_5}
\end{figure*}

\begin{figure*}[htbp]
\centerline{\includegraphics[scale=0.4]{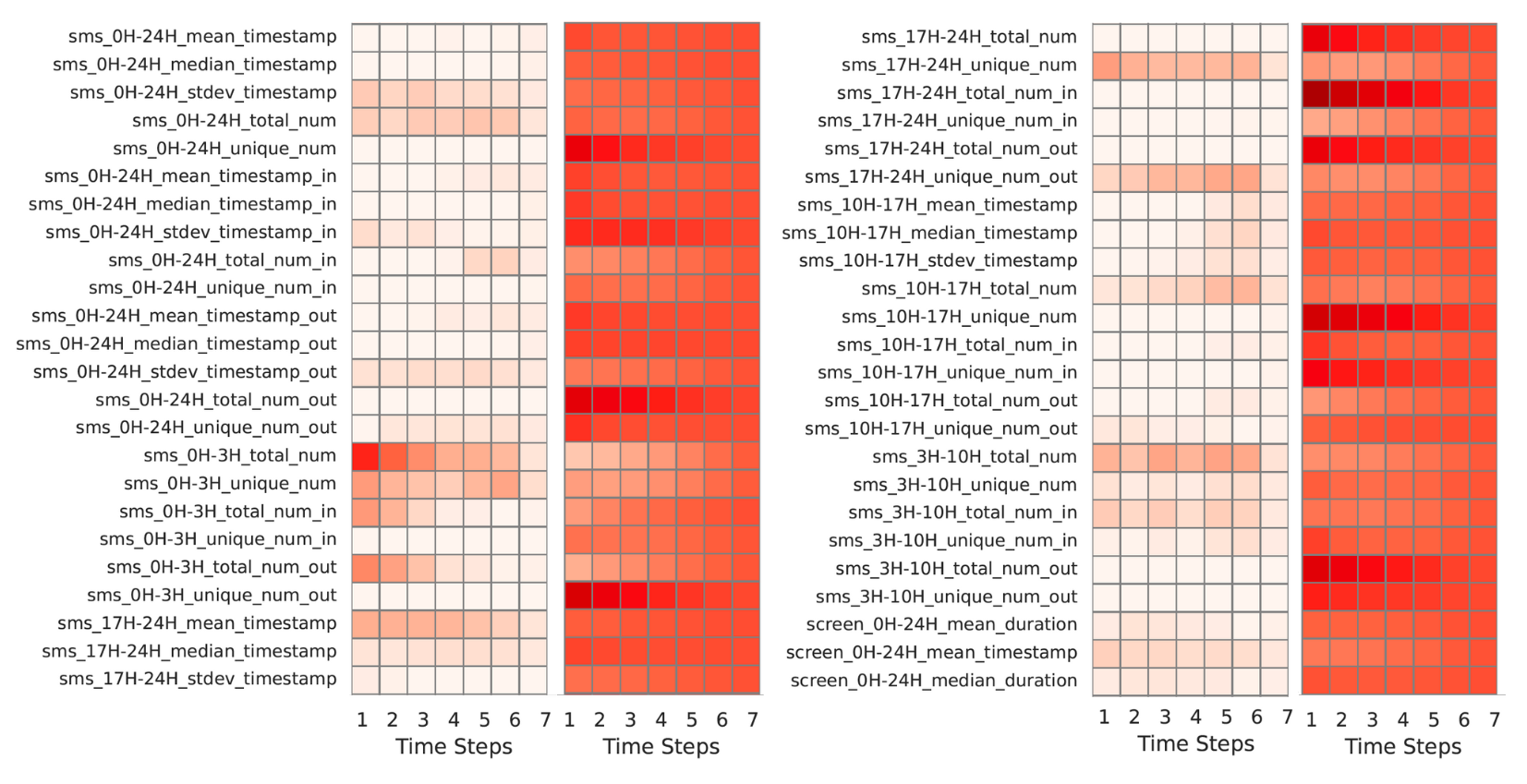}}
\caption{SNAPSHOT: Comparison of Saliency Maps From Baseline Model and Model After Proposed Method Implementation for evening-sad-happy Label Prediction for 48 features-6.}
\label{fig-snap_sal_6}
\end{figure*}

\begin{figure*}[htbp]
\centerline{\includegraphics[scale=0.4]{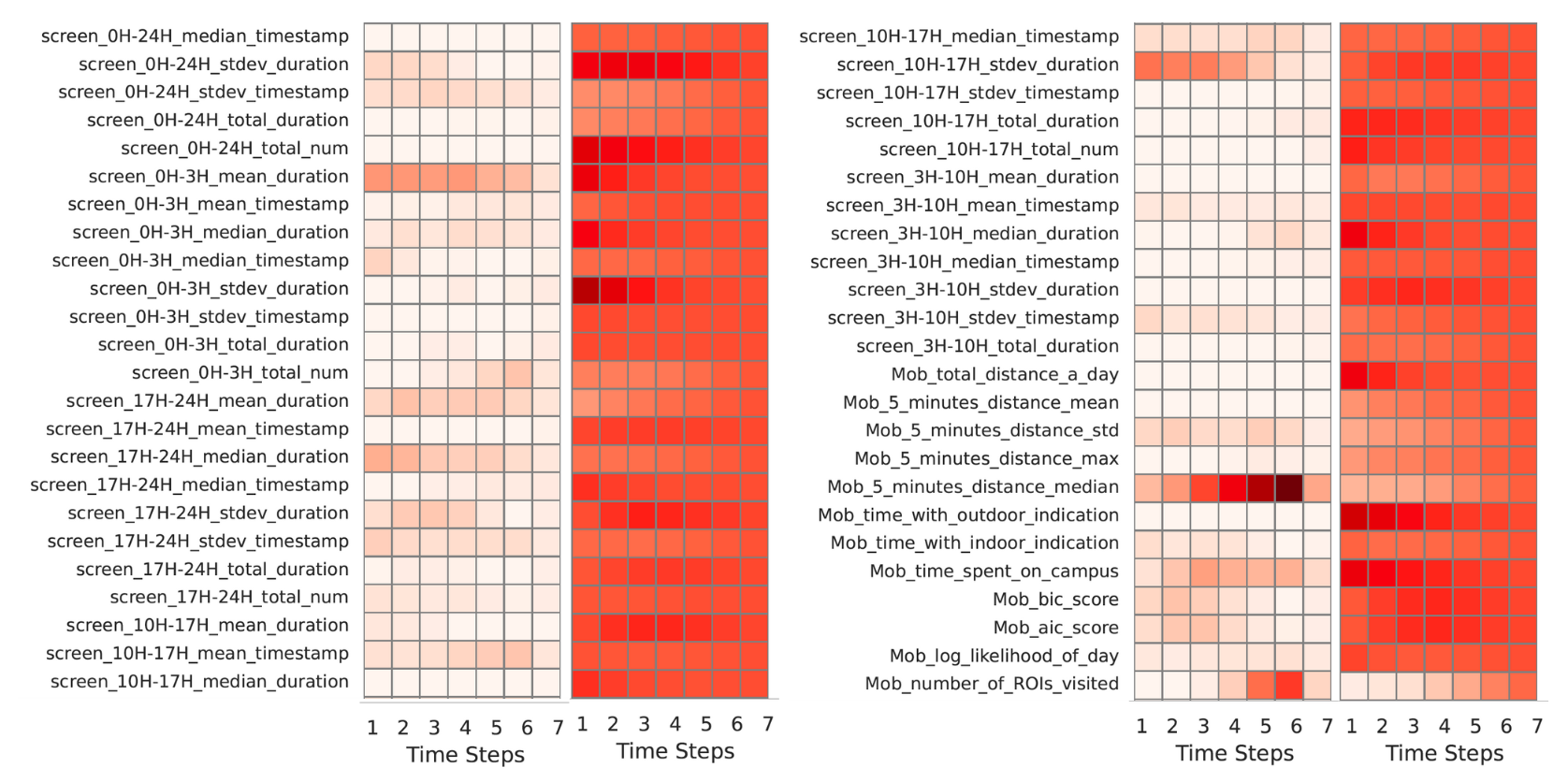}}
\caption{SNAPSHOT: Comparison of Saliency Maps From Baseline Model and Model After Proposed Method Implementation for evening-sad-happy Label Prediction for 48 features-7.}
\label{fig-snap_sal_7}
\end{figure*}

\begin{figure*}[htbp]
\centerline{\includegraphics[scale=0.4]{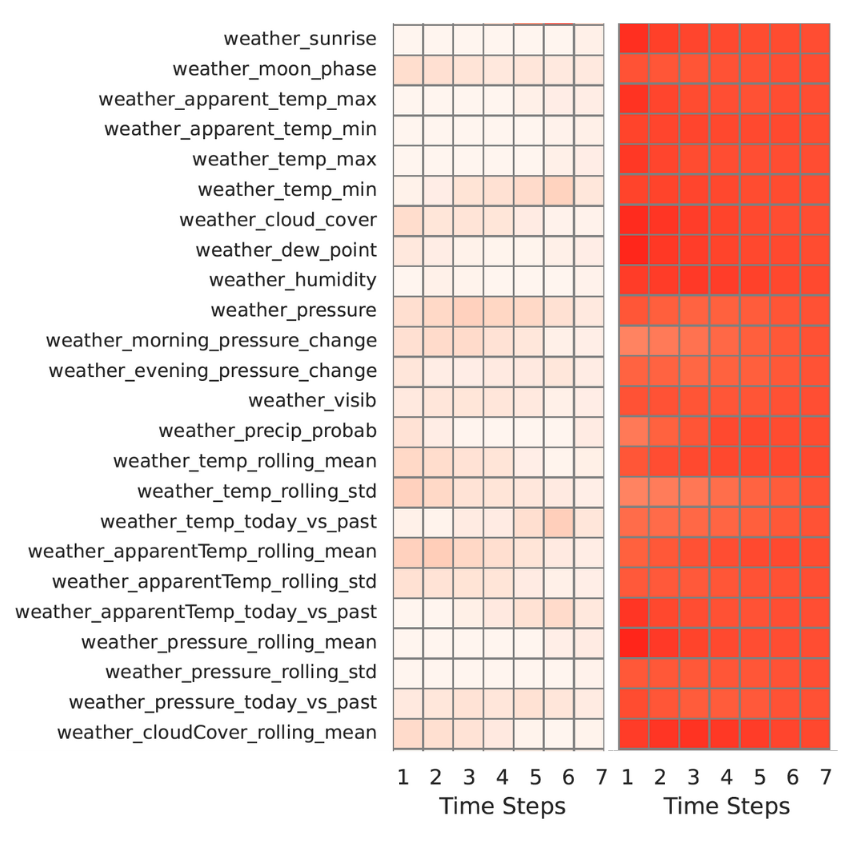}}
\caption{SNAPSHOT: Comparison of Saliency Maps From Baseline Model and Model After Proposed Method Implementation for evening-sad-happy Label Prediction for 24 features-8.}
\label{fig-snap_sal_8}
\end{figure*}

\begin{figure*}[htbp]
\centerline{\includegraphics[scale=0.4]{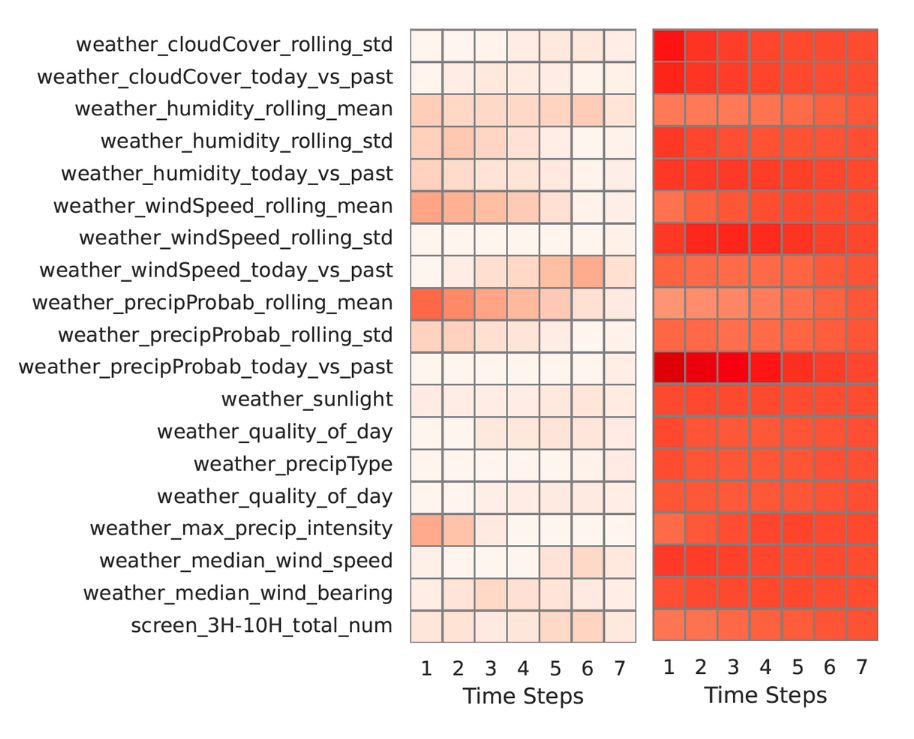}}
\caption{SNAPSHOT: Comparison of Saliency Maps From Baseline Model and Model After Proposed Method Implementation for evening-sad-happy Label Prediction for 19 features-9.}
\label{fig-snap_sal_9}
\end{figure*}

\begin{figure*}[htbp]
\centerline{\includegraphics[width=\textwidth]{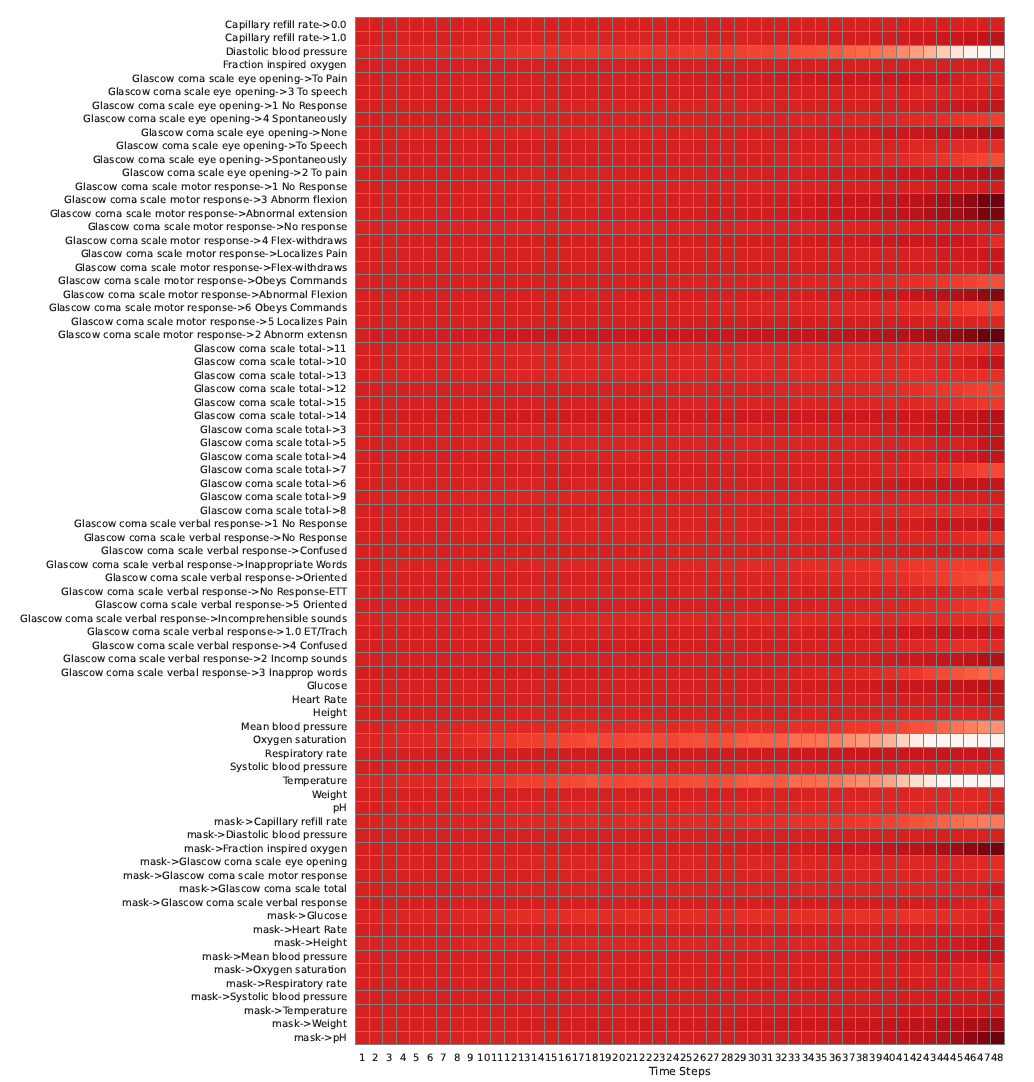}}
\caption{MIMIC-III: Saliency Maps From Baseline Model for IHM prediction.}
\label{fig-mimic_sal_baseline}
\end{figure*}

\begin{figure*}[htbp]
\centerline{\includegraphics[width=\textwidth]{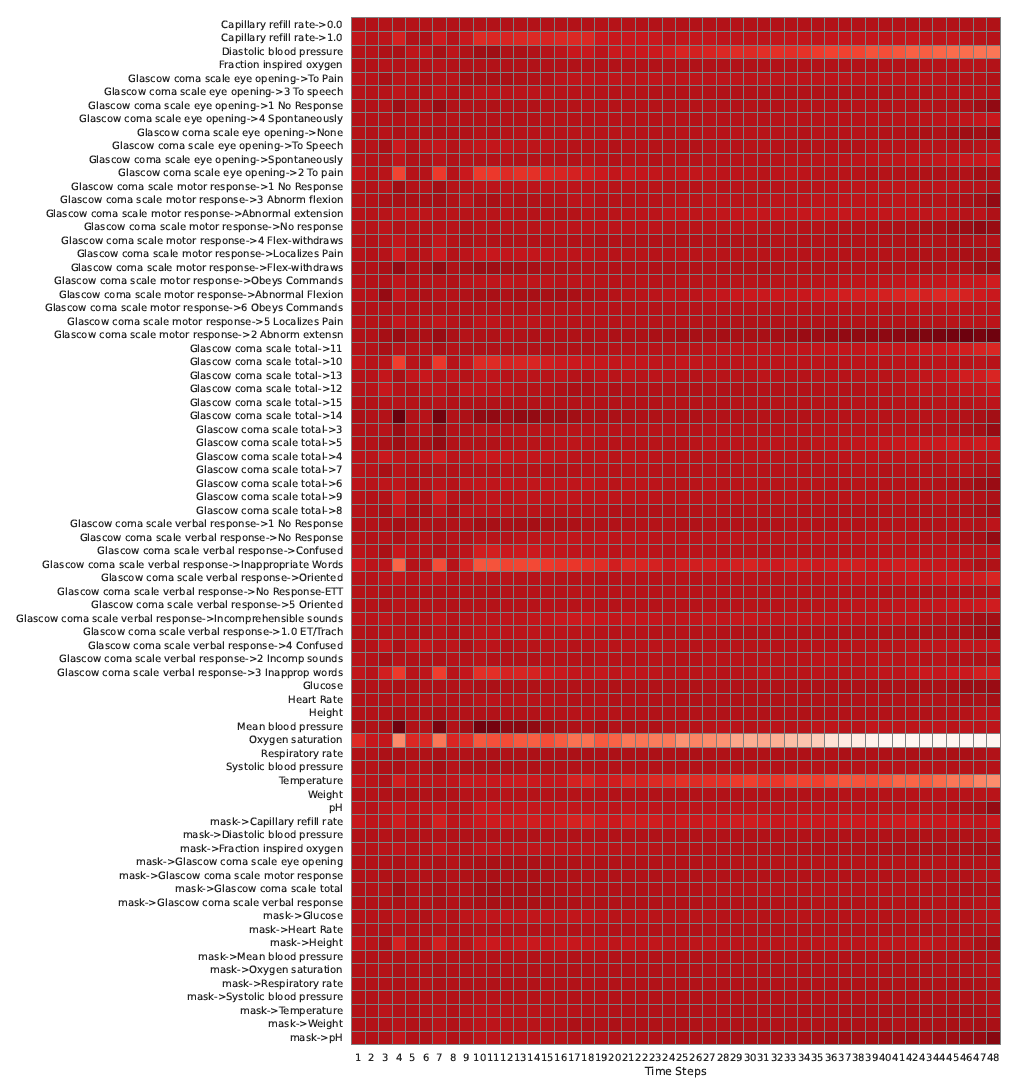}}
\caption{MIMIC-III: Saliency Maps From IHM Prediction Model after Proposed Method Implementation with Bias Mitigation by Marital Status.}
\label{fig-mimic_sal_method}
\end{figure*}

In this appendix, we provide the saliency maps for the MIMIC-III and SNAPSHOT Experiments. Figures \ref{fig-snap_sal_1} to \ref{fig-snap_sal_9} correspond with the SNAPSHOT experiments. For these figures, the images on the left represent the saliency map for the initial baseline model for evening-sad-happy prediction, and those on the right are the maps for the model generated from our proposed method. Figure \ref{fig-mimic_sal_baseline} was generated from the initial baseline model for IHM prediction for the MIMIC experiments, and Figure \ref{fig-mimic_sal_method} for the model from our proposed method implementation with bias mitigation based on marital status.

\end{document}